%% file: main.tex
\documentclass[10pt,twocolumn,letterpaper]{article}

\makeatletter
\newcommand*{\addFileDependency}[1]{% argument=file name and extension
  \typeout{(#1)}
  \@addtofilelist{#1}
  \IfFileExists{#1}{}{\typeout{No file #1.}}
}
\makeatother

\newcommand*{\myexternaldocument}[1]{%
    \externaldocument{build/#1}%
    \addFileDependency{#1.tex}%
    \addFileDependency{build/#1.aux}%
}

\usepackage[pagenumbers]{cvpr} % To force page numbers, e.g. for an arXiv version

\usepackage{graphicx}
\usepackage{amsmath}
\usepackage{amssymb}
\usepackage{booktabs}
\usepackage{xr}
\usepackage{float}
\myexternaldocument{supp}

\usepackage[pagebackref,breaklinks,colorlinks]{hyperref}

\usepackage[capitalize]{cleveref}
\crefname{section}{Sec.}{Secs.}
\Crefname{section}{Section}{Sections}
\Crefname{table}{Table}{Tables}
\crefname{table}{Tab.}{Tabs.}

\def\cvprPaperID{3766} % *** Enter the CVPR Paper ID here
\def\confName{CVPR}
\def\confYear{2023}

\begin{document}

%%%%%%%%% TITLE - PLEASE UPDATE
\title{Learning Transformations To Reduce the Geometric
Shift in Object Detection}

\author{Vidit Vidit$^1$ \  Martin Engilberge$^1$ \  Mathieu Salzmann$^{1,2}$ \\
CVLab, EPFL$^1$, ClearSpace SA$^2$\\
{\tt\small {firstname.lastname}@epfl.ch}
% For a paper whose authors are all at the same institution,
% omit the following lines up until the closing ``}''.
% Additional authors and addresses can be added with ``\and'',
% just like the second author.
% To save space, use either the email address or home page, not both
% \and
% Second Author\\
% Institution2\\
% First line of institution2 address\\
% {\tt\small secondauthor@i2.org}
}
\maketitle

\input{tex/defs}

\input{tex/abstract}
\input{tex/intro}
\input{tex/related_work}

\input{tex/approx_transform}
\input{tex/method}

\input{tex/experiments}
\input{tex/conclusion}

%%%%%%%%% REFERENCES
{\small
\bibliographystyle{ieee_fullname}
\bibliography{bibliography}
}

\end{document}

\renewcommand\thefigure{A.\arabic{figure}} 
\renewcommand\thetable{A.\arabic{table}} 

%%%%%%%%% TITLE - PLEASE UPDATE
\title{Supplementary Material: Learning Transformations To Reduce the Geometric Shift in Object Detection}

\author{Vidit Vidit$^1$ \  Martin Engilberge$^1$ \  Mathieu Salzmann$^{1,2}$ \\
CVLab, EPFL$^1$, ClearSpace SA$^2$\\
{\tt\small {firstname.lastname}@epfl.ch}
% For a paper whose authors are all at the same institution,
% omit the following lines up until the closing ``}''.
% Additional authors and addresses can be added with ``\and'',
% just like the second author.
% To save space, use either the email address or home page, not both
% \and
% Second Author\\
% Institution2\\
% First line of institution2 address\\
% {\tt\small secondauthor@i2.org}
}
\maketitle

\input{tex/defs}

\input{tex/supp}

%%%%%%%%% REFERENCES
{\small
\bibliographystyle{ieee_fullname}
\bibliography{bibliography}
}

%% file: tex/defs.tex
% !TEX root = ../main.tex
% !TEX spellcheck = en-US

\newif\ifdraft
\drafttrue

\newcommand{\bt}{\mathbf{t}}
\newcommand{\bM}{\mathbf{M}}
\newcommand{\bd}{\mathbf{d}}
\newcommand{\bX}{\mathbf{X}}
\newcommand{\bz}{\mathbf{z}}
\newcommand{\bG}{\mathbf{G}}
\newcommand{\bB}{\mathbf{B}}
\newcommand{\bS}{\mathbf{S}}
\newcommand{\bH}{\mathbf{H}}
\newcommand{\bR}{\mathbf{R}}
\newcommand{\bA}{\mathbf{A}}
\newcommand{\bdelta}{\boldsymbol{\delta}}

\newcommand{\addparagraphup}{\vspace*{0.01in}}

% corrections
\newcommand{\ms}[1]{\ifdraft {\color{green}{#1}} \else {#1}\fi}
\newcommand{\vdt}[1]{\ifdraft {\color{blue}{#1}} \else {#1}\fi}
\newcommand{\me}[1]{\ifdraft {\color{cyan}{#1}} \else {#1}\fi}

% remarks
\newcommand{\MS}[1]{\ifdraft {\color{green}{\textbf{MS: #1}}}\else {}\fi}
\newcommand{\VDT}[1]{\ifdraft {\color{red}{\textbf{VDT: #1}}}\else {}\fi}
\newcommand{\ME}[1]{\ifdraft {\color{cyan}{\textbf{ME: #1}}} \else {}\fi}
\newcommand{\chng}[1]{\ifdraft {\color{red}{#1}}\else {}\fi}

% Citations
\newcommand{\comment}[1]{}
\newcommand{\teaserfig}[1]{./images/#1}

% colors

% \iffalse
\iftrue % use space-saving macro

\newcommand{\cutsectionup}{\vspace*{0in}}
\newcommand{\cutsectiondown}{\vspace*{-0.05in}}

\newcommand{\cutsubsectionup}{\vspace*{-0.05in}}
\newcommand{\cutsubsectiondown}{\vspace*{-0.04in}}

\newcommand{\cutsubsubsectionup}{\vspace*{-0.05in}}
\newcommand{\cutsubsubsectiondown}{\vspace*{-0.05in}}

\newcommand{\cutparagraphup}{\vspace*{-0in}}
\newcommand{\cutparagraphdown}{\vspace*{-0in}}

\newcommand{\cuthalfcaptionup}{\vspace*{-0.15in}}
\newcommand{\cuthalfcaptiondown}{\vspace*{-0.25in}}

\newcommand{\cutcaptionup}{\vspace*{-0.1in}}
\newcommand{\cutcaptiondown}{\vspace*{-0.15in}}

\newcommand{\cuthalftablecaptionup}{\vspace*{-0in}}
\newcommand{\cuthalftablecaptiondown}{\vspace*{-0.1in}}

\newcommand{\cuttablecaptionup}{\vspace*{-0in}}
\newcommand{\cuttablecaptiondown}{\vspace*{-0.1in}}

\newcommand{\cutequationup}{\vspace*{-0.05in}}
\newcommand{\cutequationdown}{\vspace*{-0.01in}}

\newcommand{\cuttableup}{\vspace*{-0.2in}}
\newcommand{\cuttabledown}{\vspace*{-0.3in}}

\newcommand{\cutabstractup}{\vspace*{-0.12in}}
\newcommand{\cutabstractdown}{\vspace*{-0.3in}}

\newcommand{\cutalgorithmup}{\vspace*{-0in}}
\newcommand{\cutalgorithmdown}{\vspace*{-0.1in}}

\newcommand{\cut}{{\vspace*{-0.02in}}}
\newcommand{\cutmore}{{\vspace*{-0.06in}}}
\newcommand{\negcut}{}
\else % do not use space-saving macro
\newcommand{\cutsectionup}{}
\newcommand{\cutsectiondown}{}

\newcommand{\cutsubsectionup}{}
\newcommand{\cutsubsectiondown}{}

\newcommand{\cutsubsubsectionup}{}
\newcommand{\cutsubsubsectiondown}{}

\newcommand{\cutparagraphup}{}
\newcommand{\cutparagraphdown}{}

\newcommand{\cuthalfcaptionup}{}
\newcommand{\cuthalfcaptiondown}{}

\newcommand{\cutcaptionup}{}
\newcommand{\cutcaptiondown}{}

\newcommand{\cutequationup}{}
\newcommand{\cutequationdown}{}

\newcommand{\cuttableup}{}
\newcommand{\cuttabledown}{}

\newcommand{\cutabstractup}{}
\newcommand{\cutabstractdown}{}

\newcommand{\cutalgorithmup}{}
\newcommand{\cutalgorithmdown}{}

\newcommand{\cut}{}
\newcommand{\cutmore}{}
\newcommand{\negcut}{}
\fi

%% file: tex/abstract.tex
% !TEX root = ../main.tex
% !TEX spellcheck = en-US
\begin{abstract}
The performance of modern object detectors drops when the test distribution differs from the training one. Most of the methods that address this focus on object appearance changes caused by, e.g., different illumination conditions, or gaps between synthetic and real images. Here, by contrast, we tackle geometric shifts emerging from variations in the image capture process, or due to the constraints of the environment causing differences in the apparent geometry of the content itself\comment{the camera intrinsics or viewpoint}. We introduce a self-training approach that learns a set of geometric transformations to minimize these shifts without leveraging any labeled data in the new domain, nor any information about the cameras. We evaluate our method on two different shifts, i.e., a camera's field of view (FoV) change and a viewpoint change. Our results evidence that learning geometric transformations helps detectors to perform better in the target domains.  
\end{abstract}  

%% file: tex/intro.tex
% !TEX root = ../main.tex
% !TEX spellcheck = en-US
\section{Introduction}

While modern object detectors~\cite{ren2016faster,redmon2016you,bochkovskiy2020yolov4,glenn_jocher_2020_4154370,detr} achieve impressive results, their performance decreases when the test data depart from the training distribution. This problem arises in the presence of appearance variations due to, for example, differing illumination or weather conditions. Considering the difficulty and cost of acquiring annotated data in the test (i.e., target) domain, Unsupervised Domain Adaptation (UDA) has emerged as the standard strategy to address such scenarios~\cite{CHEN_2021_I3NET, saito2019strong,deng2021unbiased,chen2020harmonizing,zhu2019adapting}.

\input{figures/gap}

In this context, much effort has been made to learn domain invariant features, such that the source and target distributions in this feature space are similar. This has led to great progress in situations where the appearance of the objects changes drastically from one domain to the other, as in case of real-to-sketch adaptation (e.g., Pascal VOC~\cite{everingham2010pascal} to Comics~\cite{inoue2018cross}), or weather adaptation (e.g., Cityscapes~\cite{cordts2016cityscapes} to Foggy Cityscapes~\cite{sakaridis2018semantic}). Nevertheless, such object appearance changes are not the only sources of domain shifts. They can also have geometric origins. For example, as shown in \cref{fig:geoshift}, they can be due to a change in camera viewpoint or field-of-view (FoV), or a change of object scale due to different scene setups. In practice, such geometric shifts typically arise from a combination of various factors, including but not limited to the ones mentioned above.

% the bias of the cameras used and the environment, introduce shifts that change objects spatially. Additionally, such shifts are the combination of different factors, for example in the Viewpoint shift, apart from the camera orientation we change the environment from outdoor to indoor. This causes the apparent size of the bounding box (bbox) to change as it is unusual that pedestrians are directly in front of the camera in the driving car datasets.  

In this paper, we introduce a domain adaptation approach tackling such geometric shifts. To the best of our knowledge, the recent work of~\cite{gu2021pit} constitutes the only attempt at considering such geometric distortions. However, it introduces a method solely dedicated to FoV variations, assuming that the target FoV is fixed and known. Here, we develop a more general framework able to cope with a much broader family of geometric shifts.

To this end, we model geometric transformations as a combination of multiple homographies. We show both theoretically and empirically that this representation is sufficient to encompass a broad variety of complex geometric transformations. We then design an \emph{aggregator} block that can be incorporated to the detector to provide it with the capacity to tackle geometric shifts. We use this modified detector to generate pseudo labels for the target domain, which let us optimize the homographies so as to reduce the geometric shift.
%Using these pseudo-labels within a Mean Teacher~\cite{tarvainen2017mean} formalism, we optimize the homographies so as to reduce the geometric shift. 

Our contributions can be summarized as follows. (i) We tackle the problem of general geometric shifts for object detection. (ii) We learn a set of homographies using unlabeled target data, which alleviates the geometric bias arising in source-only training. (iii) Our method does not require prior information about the target geometric distortions and generalizes to a broad class of geometric shifts. Our experiments demonstrate the benefits of our approach in several scenarios. In the presence of FoV shifts, our approach yields similar performance to the FoV-dedicated framework of~\cite{gu2021pit} but without requiring any camera information. As such, it generalizes better to other FoVs. Furthermore, we show the generality of our method by using it to adapt to a new camera viewpoint in the context of pedestrian detection. 

%% file: figures/gap.tex
% !TEX root = ../main.tex
% !TEX spellcheck = en-US
\begin{figure}[h]
  \centering
  \includegraphics[width=\linewidth]{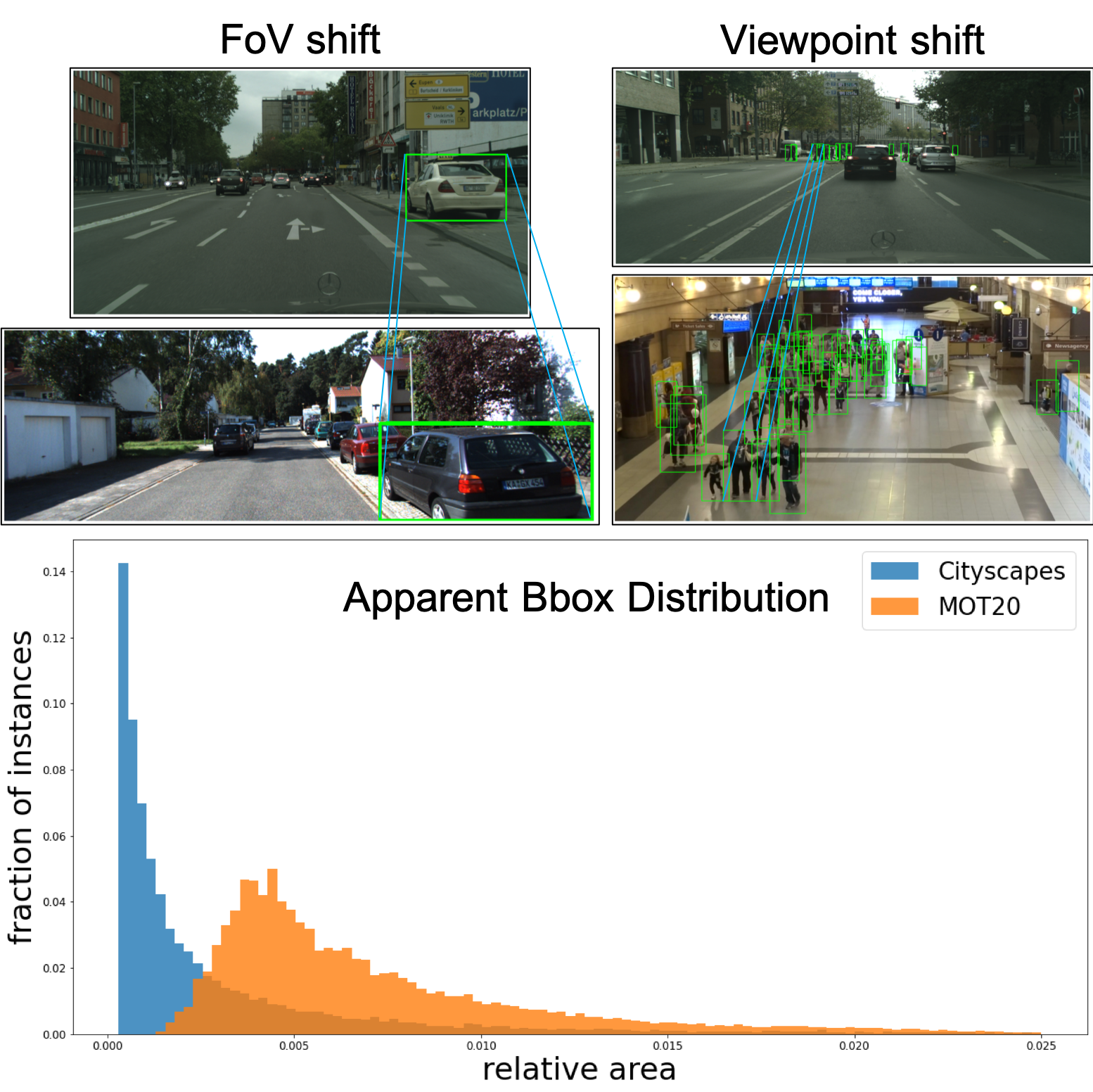}
  \caption{\textbf{Geometric shifts.} \textbf{(Left)} Due to a different FoV, the cars highlighted in green, undergo different distortions even though they appear in similar image regions. \textbf{(Right)} Different camera viewpoints (front facing vs downward facing) yield different distortions and occlusion patterns for pedestrian detection. \textbf{(Bottom)} The distributions of pedestrian bounding box sizes in Cityscapes~\cite{cordts2016cityscapes} and MOT~\cite{dendorfer2020mot20} differ significantly as the pedestrians are usually far away or in the periphery in Cityscapes. The top images are taken from Cityscapes~\cite{cordts2016cityscapes}, and the bottom-left and right ones from KITTI~\cite{geiger2012we} and MOT~\cite{dendorfer2020mot20}, respectively. }
  \label{fig:geoshift}
\end{figure}

%% file: tex/related_work.tex
\section{Related Work}

\paragraph{Unsupervised Domain Adaptation (UDA).}
UDA for image recognition~\cite{ganin2016domain,long2015learning,pei2018multi,sun2016deep,xie2018learning,tzeng2017adversarial,xu2020adversarial} and object detection ~\cite{CHEN_2021_I3NET, saito2019strong,deng2021unbiased,chen2020harmonizing,zhu2019adapting,li2022cross} has made a great progress in the past few years. The common trend in both tasks consists of learning domain invariant features. For object detection, this entails aligning the global (e.g., illumination, weather) and local (foreground objects) features in the two domains. In this context,~\cite{chen2018domain,saito2019strong,chen2020harmonizing,shen2019scl} align image- and instance-level features in the two domains via adversarial learning~\cite{ganin2016domain}; \cite{vs2021mega} learns category-specific attention maps to better align specific image regions; \cite{zhu2019adapting} clusters the proposed object regions using $k$-means clustering and uses the centroids for instance-level alignment. 
While this successfully tackles domain shifts caused by object appearance variations, it fails to account for the presence of shifts due to the image capture process itself, such as changes in camera intrinsics or viewpoint. The only initial step at considering a geometric shift is the work of~\cite{gu2021pit}, which shows the existence of an FoV gap in driving datasets~\cite{cordts2016cityscapes, geiger2012we} and proposes a Position Invariant Transform (PIT) that corrects the distortions caused specifically by an FoV change. In essence, PIT undistorts the images by assuming knowledge of the target FoV. By contrast, here, we introduce an approach that generalizes to a broad family of geometric shifts by learning transformations without requiring any camera information.

\vspace{-0.3cm}
\paragraph{Self-training.}
Self-training, generally employed in the semi-supervised setting, offers an alternative to learning domain-invariant features and utilize unlabeled data to improve a detector's performance. In this context, \cite{sohn2020simple} uses a student-teacher architecture where the teacher model is trained with supervised data and generates pseudo-labels on unannotated data. These pseudo-labels are then used to train a student model. While effective in the standard semi-supervised learning scenario, the quality of the pseudo-labels obtained with this approach tends to deteriorate when the labeled and unlabeled data present a distribution shift. \cite{deng2021unbiased,li2022cross} have therefore extended this approach to domain adaptation by using the Mean Teacher strategy of~\cite{tarvainen2017mean}  to generate reliable pseudo-labels in the target domain. Other approach include the use of CycleGAN~\cite{zhu2017unpaired} generated images to train an unbiased teacher model~\cite{deng2021unbiased}, and that of different augmentation strategies to generate robust pseudo-labels~\cite{li2022cross}.
Our approach also follows a self-training strategy but, while these works focus on object appearance shifts, we incorporate learnable blocks to address geometric shifts. As shown in our experiment, this lets us outperform the state-of-the-art AdaptTeacher~\cite{li2022cross}. 

\vspace{-0.3cm}
\paragraph{Learning Geometric Transformations.}
End-to-end learning of geometric transformations has been used to boost the performance of deep networks. For example, Spatial Transformer Networks (STNs)~\cite{jaderberg2015spatial} reduce the classification error by learning to correct for affine transformations;
%STNs can learn transformations beyond affine if the differentiabilty w.r.t. the sampled grid points is guaranteed. 
deformable convolutions~\cite{dai2017deformable}  model geometric transformations by applying the convolution kernels to non-local neighborhoods. These methods work well when  annotations are available for supervision, and make the network invariant to the specific geometric transformations seen during training. Here, by contrast, we seek to learn transformations in an unsupervised manner and allow the network to generalize to unknown target transformations.

%% file: tex/approx_transform.tex
\section{Modeling Geometric Transformations}
\label{sec:transform}

In the context of UDA, multiple geometric differences can be responsible for the gap between the domains. Some can be characterized by the camera parameters, such as a change in FoV (intrinsic) or viewpoint (extrinsic), whereas others are  content specific, such as a difference in road width between different countries.
%the geometry of a class object can differ from one domain to another. For example, the road width changes depending on the country where the data was collected. 
Ultimately, the geometric shift is typically a combination of different geometric operations. Since the parameters of these operations are unknown, we propose to bridge the domain gap by learning a geometric transform. Specifically, we aggregate the results of multiple perspective transforms, i.e., homographies, to obtain a differentiable operation that can emulate a wide variety of geometric transforms.

\subsection{Theoretical Model}
Let us first show that, given sufficiently many homographies, one can perfectly reproduce any mapping  between $\mathbb{R}^2 \setminus (0,0)$ and  $\mathbb{R}^2$.

\paragraph{Single homography for a single point.}
First, we show that a single homography with 4 degrees of freedom can map a point $p \in \mathbb{R}^2 \setminus (0,0)$ to any other point in $\mathbb{R}^2$.
To this end, let
\begin{equation}
H = \begin{bmatrix}
s_x & 0 & 0\\
0 & s_y & 0\\
l_x & l_y & 1
\end{bmatrix}
\label{eq:homography}
\end{equation}
be a homography, with $(s_x, s_y)$ the scaling factors on the $x$- and $y$-axis, respectively, and $(l_x, l_y)$ the perspective factors in $x$ and $y$, respectively.
% A point $p \in \mathbb{R}^2$ with homogeneous coordinate $[x,y,1]$ is projected to $p' = H \times p$ with homogeneous coordinate $[x',y',1]$
% Writing the projection explicitly give $x' = \frac{s_xx}{t_xx+t_yy+1}$ and $y' =  \frac{s_yy}{t_xx+t_yy+1}$.
For any destination point $d \in \mathbb{R}^2$, there exists a set of parameters $(s_x, s_y, l_x, l_y)$ such that $d = H \times p$. One such set is $(\frac{d_x}{p_x}, \frac{d_y}{p_y}, 0, 0)$.

% To be completed, multiple solution exists, should we give one? or the system of equation?

\paragraph{Emulating any geometric transformation}
Now that we have shown that a single homography can move a point to any other point in $\mathbb{R}^2$, we describe a simple protocol to emulate any geometric transform. Given an unknown geometric transform $T: \mathbb{R}^2 \setminus (0,0) \to \mathbb{R}^2$,
we aim to emulate $T$ with a set of homographies. In general, for an image $\mathbf{I} \in \mathbb{R}^{3\times h \times w}$, we can restrict the domain of $T$ to only image coordinates.
To this end, we can define a set of homographies $H_i \in \mathbb{H}$ for i in $\{1,2,3,...,h \times w\}$, where the parameters of $H_i$ are chosen to mimic the transform $T$ for location $i$ of the image. In this protocol, the aggregation mechanism is trivial since each homography is in charge of remapping a single pixel coordinate of the original space.

While this works in theory, this is of course not viable in practice since it would require too many homographies. With a smaller number of homographies, each transform needs to remap multiple points, and a more sophisticated aggregation mechanism is required. Specifically, the aggregation mechanism  needs to select which transform is in charge of remapping which point. In the next section, we empirically show that this strategy lets us closely approximate the spherical projection mapping used in PIT~\cite{gu2021pit}.

\subsection{Approximating PIT with Homographies}

To demonstrate the possibility offered by aggregating multiple homographies, we design an approximation of PIT using only homographies. PIT proposes to correct for an FoV gap by remapping images to a spherical surface. During this transformation, regions further from the center of a scene are compressed with a higher ratio. This variable compression of the space cannot be reproduced by a single homography transformation. To overcome this limitation, we combine the results of multiple homographies that all have different compression rates (scaling parameters). For the aggregation mechanism, we use the optimal strategy by selecting for each pixel the homography that approximates best the PIT mapping. As shown in \cref{fig:pit_approx_res}, this combination closely approximates the PIT results with only 5 homographies. Further analysis of these experiments is available in the supplementary material in \cref{fig:pit_approx}.

% To compare it to PIT we compute for each pixel the distance between the coordinate predicted and the one from PIT. In the top right plot of Figure~\ref{fig:pit_approx} we can see that this average per pixel coordinate error quickly goes down as we use more and more homographies. With 4 homographies the average coordinate error is already below 5 pixels, with 9 homographies it goes down to 2.5 pixels and it drops to 1.5 pixels with 25 homographies. Figure~\ref{fig:pit_approx_res} gives a qualitative results of PIT and our approximation compared to the original image.

\input{figures/pit_approx_small}

\subsection{Homographies in a Learning Setup}
In the two previous sections, we have demonstrated both theoretically and empirically the flexibility of  aggregating homographies. This makes this representation an ideal candidate for domain adaptation since the geometric shift between the domains is unknown and can be a combination of different transforms, such as FoV change, viewpoint change, camera distortion, or appearance distortion. As will be discussed in the next section, by learning jointly the set of perspective transforms and the aggregation mechanism on real data, our model can reduce the geometric shift between the two domains without prior knowledge about this domain gap.

%% file: figures/pit_approx_small.tex
\begin{figure}[h]
  \centering
  \includegraphics[width=0.8\linewidth]{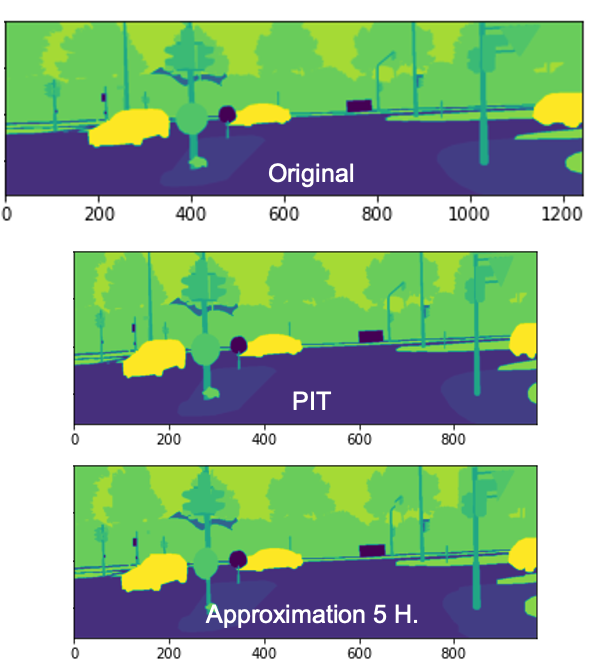}
  \caption{{\bf Approximating PIT with homographies.} We show the original image (top), the PIT~\cite{gu2021pit}  correction (middle), and our approximation of PIT using 5 homographies. Note that 5 homographies are sufficient to closely match the PIT spherical correction.}
  \label{fig:pit_approx_res}
\end{figure}

%% file: tex/method.tex
\input{figures/arch}
\section{Method}
Let us now introduce our approach to reducing the geometric shift in object detection. Following the standard UDA setting, let $D_s=\{(I_s,B_s,C_s)\}$ be a labeled source dataset containing images $I_s=\{I_s^i\}_1^{N_s}$ with corresponding object bounding boxes  $B_s=\{b_s^i\}_1^{N_s}$ and object classes $C_s=\{c_s^i\}_1^{N_s}$. Furthermore, let $D_t=\{I_t\}$ denote an unlabeled target dataset for which only images $I_t = \{I_t^i\}_1^{N_t}$ are available, without annotations. 
%We define a geometric shift between the source and target data as differences in the camera parameters used to acquire $I_s$ and $I_t$. 
Here, we tackle the case where the two domains differ by geometric shifts but assume no knowledge about the nature of these shifts.
%that these differences are unknown and may not even correspond to fixed parameters in either dataset. For example, the source data may cover a range of camera viewpoints and the target data a non-overlapping viewpoint range. 
Below, we first introduce the architecture we developed to handle this and then our strategy to train this model.

\subsection{Model Architecture}
\label{sec:method_arch}

The overall architecture of our approach is depicted in \cref{fig:arch}. In essence, and as discussed in~\cref{sec:transform}, we characterize the geometric changes between the source and target data by a set of transformations $\mathcal{T}=\{\mathcal{H}_i\}_1^N$. Each $\mathcal{H}_i$ in $\mathcal{T}$ is a homography of the same form as in~\cref{eq:homography}.
%\begin{equation}
%  \mathcal{H}_i =  \begin{bmatrix} 
%    s_x & 0 & 0 \\ 
%    0 &  s_y & 0 \\  
%    l_x & l_y & 1 
%    \end{bmatrix}\;,
%\end{equation}
%where $\{s_x,s_y\}$ represent scaling factors and $\{l_x,l_y\}$ perspective effects in the $x$ and $y$ image directions. 
For our method to remain general, we assume the transformations to be unknown, and our goal, therefore, is to learn $\mathcal{T}$ to bridge the gap between the domains. This requires differentiability w.r.t. the transformation parameters, which we achieve using the sampling strategy proposed in~\cite{jaderberg2015spatial}. 

As shown in \cref{fig:arch}, the input image is transformed by the individual homographies in $\mathcal{T}$, and the transformed images are fed to a modified FasterRCNN~\cite{ren2016faster} detector. Specifically, we extract a feature map $\mathcal{F}_{\mathcal{H}_i}\in \mathbb{R}^{H\times W \times C}$ for each transformed image via a feature extractor shared by all transformations. To enforce spatial correspondence between the different $\mathcal{F}_{\mathcal{H}_i}$s, we unwarp them with $\mathcal{H}_i^{-1}$. 

We then introduce an \emph{aggregator} $\mathcal{A}_{\theta_g}$, parameterized by $\theta_g$, whose goal is to learn a common representation given a fixed number of unwarped feature maps $\mathcal{F}_{\mathcal{H}_i}^{\prime}$. To achieve this, the aggregator takes as input
\begin{equation}
    \mathcal{G} = \mathcal{F}_{\mathcal{H}_1}^{\prime} \oplus \mathcal{F}_{\mathcal{H}_2}^{\prime} \oplus ... \oplus \mathcal{F}_{\mathcal{H}_N}^{\prime} \in \mathbb{R}^{H\times W \times C\times N}\;,
\end{equation}
where $\oplus$ represents concatenation in the channel dimension. The aggregator outputs a feature map $\mathcal{A}_{\theta_g}(\mathcal{G})\in \mathbb{R}^{H\times W\times C}$, whose dimension is independent of the number of transformations. This output is then passed to a detection head to obtain the objects' bounding boxes and class labels.

\subsection{Model Training}
\label{sec:method_training}
Our training procedure relies on three steps: (i) Following common practice in UDA, we first train the FasterRCNN detector with source-only data; (ii) We then introduce the aggregator and train it so that it learns to combine different homographies using the labeled source data; (iii) Finally, we learn the optimal transformations for adaptation using both the source and target data via a Mean Teacher~\cite{tarvainen2017mean} strategy.

\paragraph{Aggregator Training.} To train the aggregator, we randomly sample a set of homographies $\mathcal{T}\in \mathbb{R}^{N\times 4}$ in each training iteration.\footnote{As our homographies involve only 4 parameters, with a slight abuse of notation, we say that $\mathcal{H}_i\in \mathbb{R}^4$.} This gives the aggregator the ability to robustly combine diverse input transformations but requires strong supervision to avoid training instabilities. We, therefore, perform this step using the source data.

The loss function for a set of transformed images $\mathcal{T}(I_s)$ is then defined as in standard FasterRCNN training with a combination of classification and regression terms~\cite{ren2016faster}. That is, we train the aggregator by solving
\begin{equation}
    \min_{\theta_{g}} \mathcal{L}_{cls}(\mathcal{T}(I_s))+\mathcal{L}_{reg}(\mathcal{T}(I_s))\;,
\end{equation}
where
\begin{align}
    \mathcal{L}_{cls}(\mathcal{T}(I_s)) &= \mathcal{L}_{cls}^{rpn} + \mathcal{L}_{cls}^{roi}\;, \\
    \mathcal{L}_{reg}(\mathcal{T}(I_s)) &= \mathcal{L}_{reg}^{rpn} + \mathcal{L}_{reg}^{roi}\;.
\end{align}
$\mathcal{L}^{rpn}_{\cdot}$ and $\mathcal{L}^{roi}_{\cdot}$ correspond to  the Region Proposal Network (RPN) loss terms and the Region of Interest (RoI) ones, respectively. 
During this process, we freeze the parameters $\theta_{b}$ of the \emph{base} network, i.e, feature extractor and detection head, which were first trained on the source data without aggregator. %This makes our method different than standard data augmentation where the feature extractor is trained with several transformations. 
Ultimately, the aggregator provides the network with the capacity to encode different transformations that are not seen in the source domain. The third training step then aims to learn the best transformation for successful object detection in the target domain.
%Moreover, since we can use STNs~\cite{jaderberg2015spatial} sampler for projection, $\mathcal{T}$ can be optimised w.r.t detector loss.

\paragraph{Learning the Transformations.}
As we have no annotations in the target domain, we exploit a Mean Teacher (MT) strategy to learn the optimal transformations. To this end, our starting point is the detector with a trained aggregator and a set of random transformations $\mathcal{T}$. The MT strategy is illustrated in \cref{fig:meanteacher}. In essence, MT training~\cite{tarvainen2017mean} involves two copies of the model: A student model, with parameters $\theta^{st}=\{\mathcal{T}^{st},\theta_b^{st},\theta_g^{st}\}$, that will be used during inference, and a teacher model, with parameters $\theta^{te}=\{\mathcal{T}^{te},\theta_b^{te},\theta_g^{te}\}$, that is updated as an Exponentially Moving Average (EMA) of the student model. That is, the student's parameters are computed with standard backpropagation, whereas the teacher's ones are updated as
\begin{equation}
    \theta_{te} \leftarrow \alpha \theta_{te} + (1-\alpha) \theta_{st}\;.
\end{equation}

The student model is trained using both source and target detection losses. Since the target domain does not have annotations, the teacher model is used to generate pseudo-labels. These labels might be noisy, and hence we only keep the predictions with a confidence score above a threshold $\tau$. Furthermore, non-maxima suppression (NMS) is used to remove the highly-overlapping bounding box predictions. 

\input{figures/meanteacher}

Formally, given a source image $I_s$ and a target image $I_t$, the student model is trained by solving
\begin{equation}
    \min_{\mathcal{T}^{st},\theta_{g}^{st},\theta_{b}^{st}} \mathcal{L}_{det}(\mathcal{T}(I_s)) + \lambda \mathcal{L}_{det}(\mathcal{T}(I_t))\;,
\end{equation}
where $\lambda$ controls the target domain contribution and
\begin{align}
    \mathcal{L}_{det}(\mathcal{T}(I_s)) &= \mathcal{L}_{cls}(\mathcal{T}(I_s))+\mathcal{L}_{reg}(\mathcal{T}(I_s))\;, \label{eq:detsource} \\
    \mathcal{L}_{det}(\mathcal{T}(I_t)) &= \mathcal{L}_{cls} (\mathcal{T}(I_t)) \;. \label{eq:dettarget} 
\end{align}
Similarly to~\cite{li2022cross,kim2019self}, we update the student model with only the classification loss in the target domain to help stabilize training.

%% file: figures/arch.tex
% !TEX root = ../main.tex
% !TEX spellcheck = en-US
\begin{figure*}
  \centering
  \includegraphics[width=\textwidth]{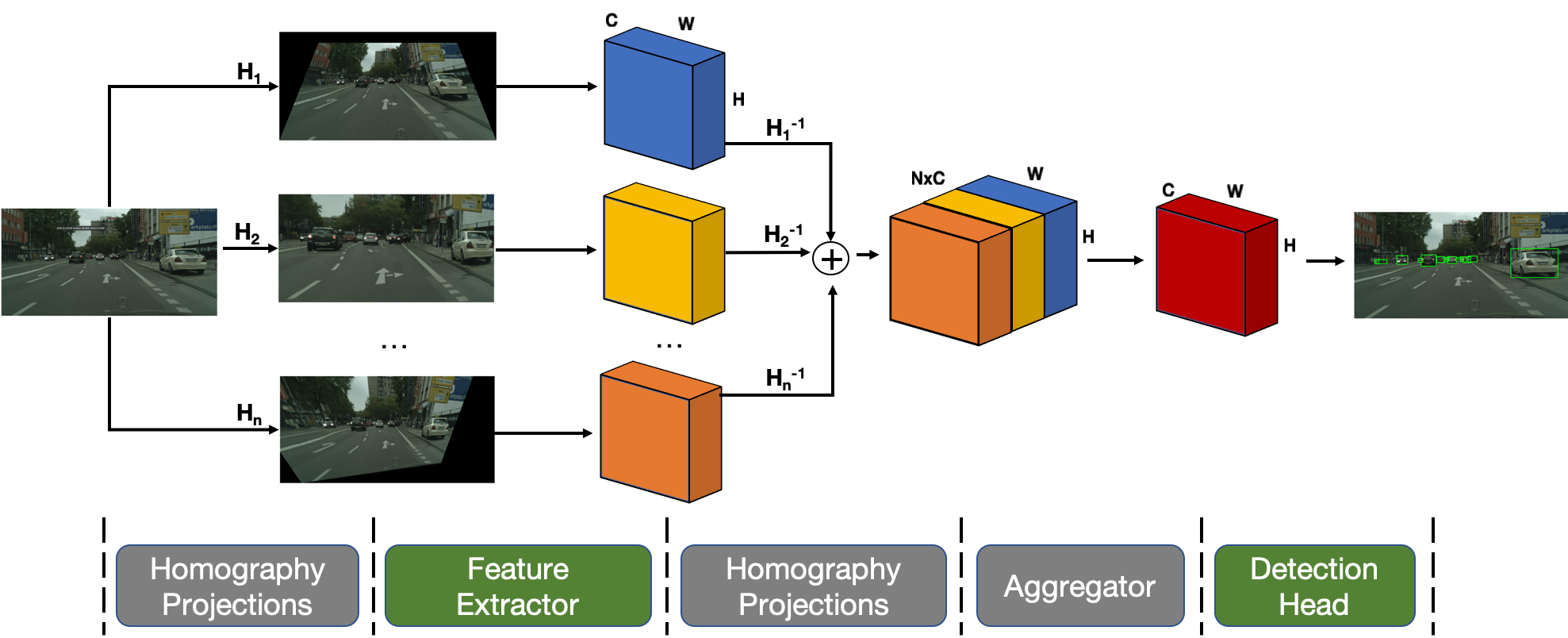}
  \caption{\textbf{Architecture}: The input image is first transformed by a set of trainable homographies. The feature maps extracted from the transformed images are then unwarped by the inverse homographies to achieve spatial consistency. We then combine the unwarped feature maps using a trainable \emph{aggregator}, whose output is passed to a detection head. The blocks shown in \textcolor[rgb]{0,0.25,0}{green} correspond to standard FasterRCNN operations. The $\bigoplus$ symbol represents the concatenation operation.}
  \label{fig:arch}
\end{figure*}

%% file: figures/meanteacher.tex
% !TEX root = ../main.tex
% !TEX spellcheck = en-US
\begin{figure}[t]
  \centering
  \includegraphics[width=\linewidth]{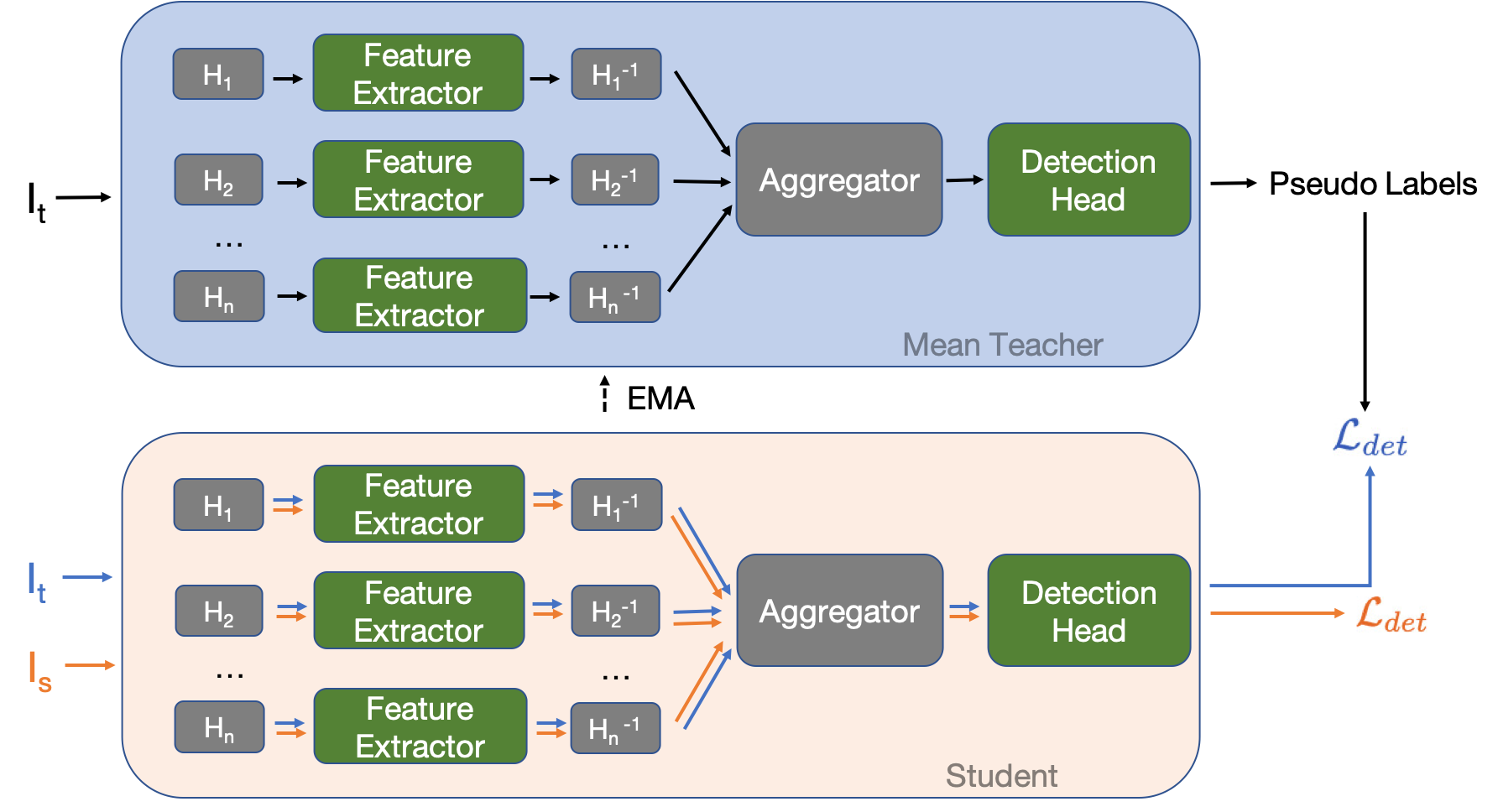}
  \caption{{\bf Mean Teacher formalism.} The student model is trained with ground-truth labels in the source domain and pseudo labels in the target one. These pseudo labels are produced by the teacher model, which corresponds to an exponentially moving average (EMA) of the student network. }
  \label{fig:meanteacher}
\end{figure}

%% file: tex/experiments.tex
\section{Experiments}
We demonstrate the effectiveness and generality of our method on different geometric shifts. First, to compare to the only other work that modeled a geometric shift~\cite{gu2021pit}, we tackle the problem of a change in FoV between the source and target domain. Note that, in contrast to~\cite{gu2021pit}, we do not assume knowledge of the target FoV. Furthermore, while~\cite{gu2021pit} was dedicated to FoV adaptation, our approach generalizes to other geometric shifts. We demonstrate this on the task of pedestrian detection under a viewpoint shift. We compare our method with the state-of-the-art AdaptTeacher~\cite{li2022cross}, which also uses a Mean Teacher, but focuses on appearance shifts. In the remainder of this section, we describe our experimental setup and discuss our results.

\subsection{Datasets}
\label{sec:expe_datasets}
\paragraph{Cityscapes~\cite{cordts2016cityscapes}} contains 2975 training and 500 test images with annotations provided for 8 categories (\emph{person}, \emph{car}, \emph{train}, \emph{rider}, \emph{truck}, \emph{motorcycle},
\emph{bicycle} and \emph{bus}). 
%These are street-view images taken from a car. 
The average horizontal (FoV\emph{x}) and vertical (FoV\emph{y}) FoVs of the capturing cameras are 50\textdegree and 26\textdegree, respectively. We use this dataset as the source domain for both FoV adaptation and viewpoint adaptation.

\paragraph{KITTI~\cite{geiger2012we}} is also a street-view dataset containing 6684 images annotated with the \emph{car} category. The horizontal (FoV\emph{x}) and vertical (FoV\emph{y}) FoVs of the camera are 90\textdegree and 34\textdegree, respectively. We use this dataset as target domain for FoV adaptation, as the viewpoint is similar to that of Cityscapes. Following~\cite{gu2021pit}, we use 5684 images for unsupervised training and 1000 images for evaluation.

\paragraph{MOT~\cite{dendorfer2020mot20}}  is a multi-object tracking dataset. We use the indoor mall sequence, MOT20-02, consisting of 2782 frames annotated with the \emph{person} category. We employ this dataset as target domain for viewpoint adaptation. We use the first 2000 frame for unsupervised training and last 782 for evaluation.

\subsection{Adaptation Tasks and Metric}
\paragraph{FoV adaptation.} As in~\cite{gu2021pit}, we consider the case of an increasing FoV using Cityscapes as source domain and KITTI as target domain. 
The horizontal and vertical FoVs increase from (50\textdegree, 26\textdegree) in Cityscapes to (90\textdegree, 34\textdegree) in KITTI. Therefore, as can be seen in \cref{fig:geoshift}, the KITTI images have a higher distortion in the corners than the Cityscapes ones. Similarly to PIT~\cite{gu2021pit}, we use the \emph{car} category in our experiments.

\paragraph{FoV generalization.} Following PIT~\cite{gu2021pit}, we study the generalization of our approach to new FoVs by cropping the KITTI images to mimic different FoV changes in the horizontal direction (FoV\emph{x}). Specifically, we treat FoV\emph{x} = 50\textdegree \,as the source domain and the cropped images with FoV\emph{x} = \{70\textdegree, 80\textdegree, 90\textdegree\} \,as different target domains. We evaluate our approach on \emph{car} on these different pairs of domains.

\paragraph{Viewpoint adaptation.} This task entails detecting objects seen from a different viewpoint in the source and target domains. We use the front-facing Cityscapes images as source domain and the downward-facing MOT ones as target one. As the MOT data depicts  pedestrians, we use the bounding boxes corresponding to the \emph{person} category in Cityscapes.\footnote{In Cityscapes, a person may be labeled as either \emph{person} or \emph{rider}. Since the \emph{rider} label is used for people riding a vehicle, we omit these cases.}

\paragraph{Metric.} In all of our experiments, we use the Average Precision (AP) as our metric. Specifically, following~\cite{gu2021pit}, we report the AP@0.5, which considers the predictions as true positives if they  match the ground-truth label and have an intersection over union (IOU) score of more than 0.5 with the ground-truth bounding boxes.

\subsection{Implementation Details}
\label{sec:expe_implementation}
We use the Detectron2~\cite{wu2019detectron2} implementation of FasterRCNN~\cite{ren2016faster} with a ResNet50~\cite{he2016deep} backbone as our \emph{base} architecture. In all of our experiments, the images are resized so that the shorter side has 800 pixels while maintaining the aspect ratio. The base network is first trained on source-only images with random cropping and random flipping augmentation for 24k iterations with batch size 8. We use the Stochastic Gradient Descent (SGD) optimizer with a learning rate of 0.01, scaled down by a 0.1 factor after 18k iterations. We use ImageNet~\cite{ILSVRC15} pretrained weights to initialize the ResNet50 backbone.

We then incorporate the \emph{aggregator} in the trained base architecture.
The aggregator architecture contains three convolutional layers with a kernel size of $3\times 3$, and one $1\times 1$ convolutional layer. We first train the aggregator on the source data with the base frozen and using random transformations $\mathcal{T}$. The transformations are generated by randomly sampling each $\mathcal{H}_i$ parameters as $s_x,s_y \sim \mathcal{U}_{[0.5,2.0]},\mathcal{U}_{[0.5,2.0]}$ and $l_x,l_y \sim \mathcal{U}_{[-0.5,0.5]},\mathcal{U}_{[-0.5,0.5]}$. 
We train the aggregator for 30k iterations using a batch size of 8 and the SGD optimizer with a learning rate of $1e^{-4}$.

The student and teacher models are then initialized with this detector and the random $\mathcal{T} = \{\mathcal{H}_i\}_{i=1}^N$. We optimize $\mathcal{T}$ using Adam~\cite{kingma2014adam}, while the base and aggregator networks are optimized by SGD. The learning rate is set to $1e^{-3}$ and scaled down by a factor 0.1 after 10k iterations for the SGD optimizer. For the first 10k iterations in FoV adaptation and for 2k iterations for viewpoint adaptation, we only train $\mathcal{T}$ keeping base and aggregator frozen. The $\alpha$ coefficient for the EMA update is set to $0.99$; the confidence threshold $\tau= 0.6 $;  $\lambda=\{0.01,0.1\}$ for FoV and viewpoint adaptation, respectively. The Mean Teacher framework is trained using both the source and target data. We set $N=5$, unless otherwise specified, and use a batch size of 4, containing 2 source and 2 target images. We apply random color jittering on both the source and target data as in~\cite{tarvainen2017mean,li2022cross}. All of our models are trained on a single NVIDIA V100 GPU. A detailed hyper-parameter study is provided in the supplementary material.

\subsection{Comparison with the State of the Art}
\label{sec:exp_sota}

\input{figures/sota_kitty_visu}

We compare our approach with the following baselines. \textbf{FR}: FasterRCNN trained only on the source data with random crop augmentation; \textbf{AT}: AdaptTeacher~\cite{li2022cross}; \textbf{MT}: Mean Teacher initialized with FR and trained with random color jittering on both the source and target data (i.e., this corresponds to our mean teacher setup in \cref{sec:method_training} but without the aggregator and without transformations $\mathcal{T}$); \textbf{FR+PIT}: Same setup as FR but with the images corrected with PIT~\cite{gu2021pit}; \textbf{MT+PIT}: Same setup as MT but with the images corrected with PIT. We refer to our complete approach (\cref{sec:method_training}) as \textbf{Ours}. For the task of FoV generalization, we report our results as \textbf{Ours-\emph{h}} to indicate that we only optimize the homographies ($5\times 4$ parameters) in $\mathcal{T}$ to adapt to the new FoVs while keeping  the base and aggregator networks frozen. This matches the setup of  PIT~\cite{gu2021pit}, which also corrects the images according to the new FoVs. As \textbf{Ours} and \textbf{Ours-\emph{h}} are trained with randomly initialized $\mathcal{T}$, we report the average results and standard deviations over three independent runs.

% \textbf{Ours-\emph{h}}: during our mean teacher training, we only optimize $\mathcal{T}$ while keeping \emph{base} and \emph{aggregator} frozen; \textbf{Ours}: our full step up detailed in section~\ref{sec:method_training}
\input{tables/fovcity2kitty}

\input{tables/fovgeneralise}

\vspace{-0.5em}
\paragraph{FoV adaptation.} The results of Cityscapes $\rightarrow$ KITTI FoV adaptation are provided in \cref{tab:city2kittyfov}. Both MT+PIT and \emph{Ours} both bridge the FoV gap, outperforming the MT baseline. Note, however, that we achieve this by learning the transformations, without requiring any camera-specific information, which is needed by PIT. Note also that MT outperforms FR by learning a better representation in the target domain, even though FR is trained with strong augmentation, such as random cropping. AT underperforms because its strong augmentation strategy fails to generalize for datasets having prominent geometric shifts. Our improvement over MT evidences that learning transformations helps to overcome geometric shifts. We optimize with $N=9$, homographies in this setup. \cref{fig:sota_visu_kitty} shows a qualitative example. Different homographies look into different image regions and the aggregator learns how to combine the activations corresponding to objects as depicted in ~\cref{fig:tfactivations_small}.

\paragraph{FoV generalization.}~\cref{tab:fovgeneralize} summarizes the results obtained by using different FoVs as target domains while fixing the source FoV to 50\textdegree. Since both the source and target images are taken from KITTI, the domain gap is only caused by a FoV change. Note that the performance of FR drops quickly as the FoV gap increases. \emph{Ours-h} outperforms FR+PIT by a growing margin as the FoV gap increases. This shows that learning transformations helps to generalize better to different amounts of geometric shifts.

\paragraph{Viewpoint adaptation.} As shown in \cref{fig:geoshift}, a change in the camera viewpoint yields differences in the observed distortions and type of occlusions. The results in \cref{tab:city2mot} show the benefits of our method over MT in this case. Note that PIT, which was designed for FoV changes, cannot be applied to correct for a viewpoint change. Other baselines outperform FR, as they use pseudo labels to fix the difference in bounding box distribution, as shown in ~\cref{fig:geoshift}. 
%AT underperforms compared to our method as there is a limited domain shift due to only appearance change.
These results illustrate the generality of our method to different kinds of geometric shifts. Qualitative results for this task can be found in~\cref{fig:sota_visu_mot}.
% shows a qualitative example of this task.

\input{tables/mot}

\input{figures/featact_small.tex}
\subsection{Additional Analyses}
\label{sec:expe_additioanl}
\paragraph{Variable number of homographies.}
\input{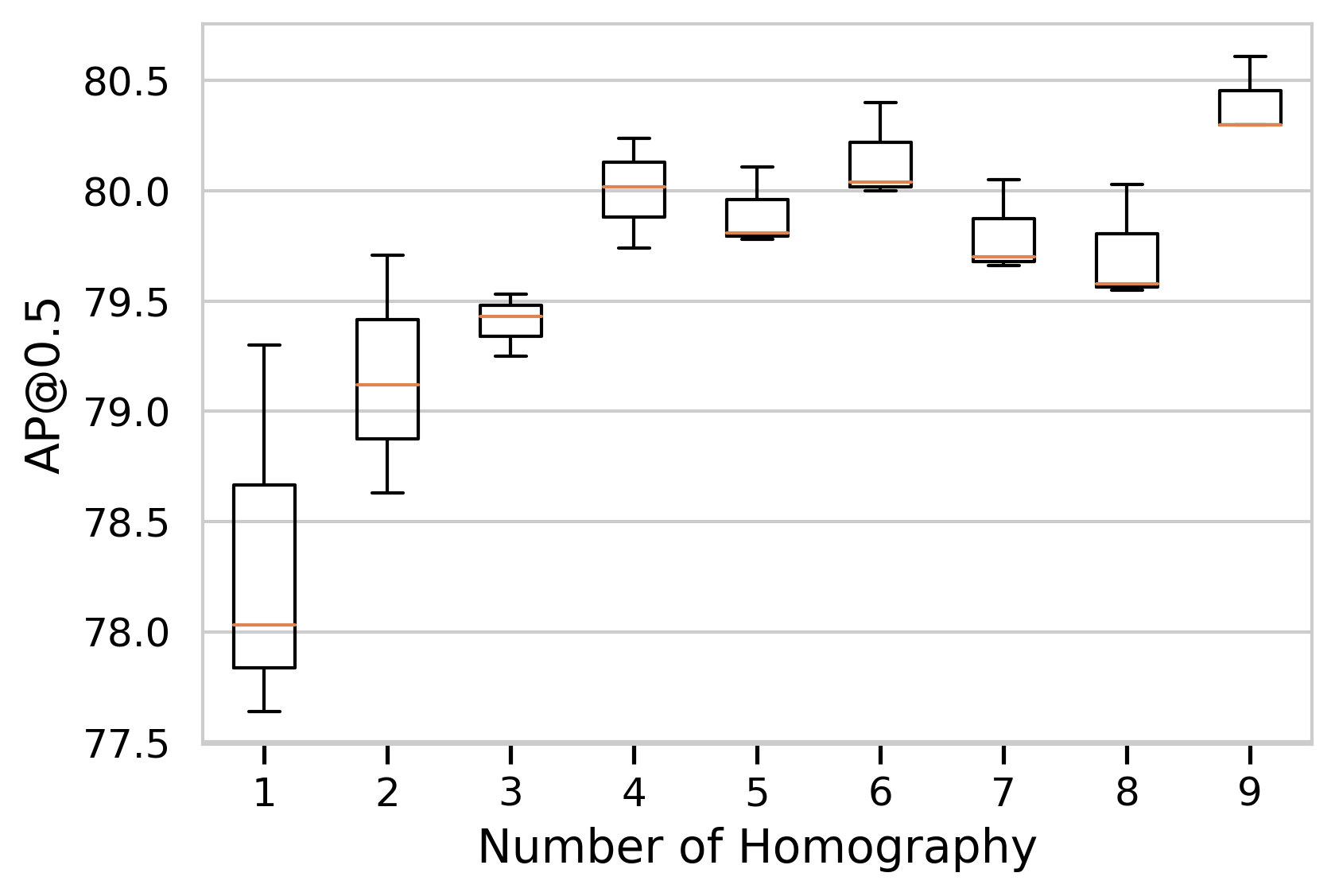}
Let us now study the influence of the number of homographies in $\mathcal{T}$. To this end, we vary this number between 1 and 9. In \cref{fig:nstn}, we plot the resulting APs for the Cityscapes-to-KITTI FoV adaptation task. Increasing the number of transformations results in a steady increase in performance, which nonetheless tends to plateau starting at 4 homographies. Due to limited compute resources, we couldn't run experiments with more than 9 homographies. This confirms the intuition that a higher number of perspective transformations can better capture the geometric shift between two domains. Therefore, we conducted all experiments with the maximum number of homographies allowed by our compute resources. 

\paragraph{Only optimizing $\mathcal{T}$.} We also run the \emph{Ours-h} baseline in the FoV and viewpoint adaptation scenarios. The resulting APs are 78.2 and 49.8, respectively. By learning only the 20 ($5\times 4$) homography parameters, our approach outperforms FR (in \cref{tab:city2kittyfov} and \cref{tab:city2mot}, respectively) by a large margin in both cases. This confirms that our training strategy is able to efficiently optimize $\mathcal{T}$ to bridge the geometric gap between different domains. We visualize in \cref{fig:evolution} in the supplementary material some transformations learned for FoV adaptation by \emph{Ours-h}. Note that they converge to diverse homographies that mimic a different FoV, correctly reflecting the adaptation task.

\paragraph{Diversity in $\mathcal{T}$.} To show that our approach can learn a diverse set of transformations that help in the adaptation task, we initialize all the homographies with identity.
%and train with our approach.
\cref{fig:eval_id_kitti_main} depicts the diversity of the learned homographies on the FoV adaptation task. Even though we do not enforce diversity, our approach learns a diverse set of transformations. With these learned homorgraphies, our model achieves 79.5 AP@0.5 score for the FoV adaptation task. We show additional results in the supplementary material~\cref{sec:diversityinT} and~\cref{sec:evolutionofT}.

\input{figures/evol_from_identity_main.tex}

\paragraph{Limitations.}
%\paragraph{Generic perspective transformations} 
Our approach assumes that the geometric gap between two domains can be bridged by a set of perspective transformations. We have shown that with enough transformations this is true. However, using a large number of homographies comes at a computational cost. The computational overhead leads to an increment in the inference time from $0.062s$ to $0.096$s for $N=5$ on an A100 Nvidia GPU with image dimension $402\times 1333$. Nevertheless, our simple
implementation shows promising results, and we will work on reducing this overhead in future work.
%both in terms of time and memory. 
Moreover since the optimization of the homography set is done at the dataset level, only certain transformations are beneficial to a given image. In the future, we therefore intend to condition the homography on the input image, which would reduce the total number of homographies needed.

%% file: figures/sota_kitty_visu.tex
% !TEX root = ../main.tex
% !TEX spellcheck = en-US
\begin{figure}
  \centering
  \includegraphics[width=\linewidth]{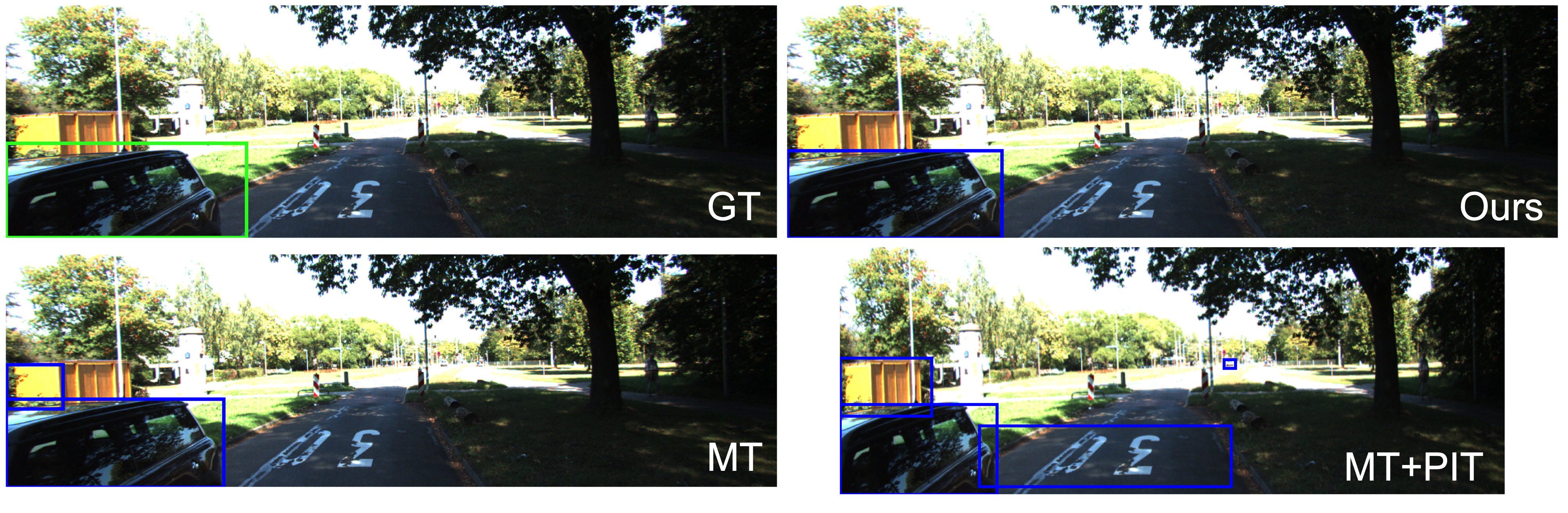}
  \caption{\textbf{FoV Adaptation: Qualitative Results.} We visualize a car detection result in the Cityscapes-to-KITTI FoV adaptation scenario. The top left image corresponds to the ground truth, the bottom left to the Mean Teacher result, which confuses the orange container with a car, the bottom right to the Mean Teacher adaptation + PIT FoV adaptation result, which also mistakes the orange container for a car and further detects the speed limit on the road. Our approach, on the top right, correctly matches the ground truth.}
  \label{fig:sota_visu_kitty}
\end{figure}

%% file: tables/fovcity2kitty.tex
% !TEX root = ../main.tex
% !TEX spellcheck = en-US
\begin{table}
% \begin{minipage}{.3\textwidth}
  \centering
  \begin{tabular}{lc}
    \toprule
    Method     & Car AP@0.5      \\
    \midrule
    FR~\cite{ren2016faster} & 76.1\\
    AT~\cite{li2022cross} & 77.2 \\
    FR+PIT & 77.6\\
    MT  & 78.3\\
    MT+PIT~\cite{gu2021pit} & 79.7 \\
   % Ours-\emph{h} & 78.2\\
    \textbf{Ours}     &  \textbf{80.4} \tiny{$\pm$ 0.15}     \\
    \bottomrule
  \end{tabular}
  \caption{FoV Adaptation.}
  \label{tab:city2kittyfov}
% \end{minipage}%
% \begin{minipage}{.75\textwidth}
%   \centering
%   \begin{tabular}{lcccc}
%     \toprule
%     \multicolumn{5}{c}{ Car AP@0.5 for FoV\emph{x}}                   \\
%     \cmidrule(r){2-5}
%     Method     & 50\textdegree & 70\textdegree & 80\textdegree & 90\textdegree       \\
%     \midrule
%     FR~\cite{ren2016faster} & 94.3 & 90.2  & 86.8 & 80.6  \\
%     FR+PIT~\cite{gu2021pit} & 93.6 & 91.4 & 89.2 & 85.9     \\
    
%     \textbf{Ours-\emph{h}} & \textbf{94.1}\tiny{$\pm$ 0.16} & \textbf{93.1}\tiny{$\pm$ 0.33} & \textbf{91.8}\tiny{$\pm$ 0.40} & \textbf{88.8}\tiny{$\pm$ 0.21} \\
%     \bottomrule
%   \end{tabular}
%   \caption{FoV Generalization}
%   \label{tab:fovgeneralize}
% \end{minipage}

\end{table}

%% file: tables/fovgeneralise.tex
% !TEX root = ../main.tex
% !TEX spellcheck = en-US
\begin{table}

  \centering
  \begin{tabular}{lcccc}
    \toprule
    \multicolumn{5}{c}{ Car AP@0.5 for FoV\emph{x}}                   \\
    \cmidrule(r){2-5}
    Method     & 50\textdegree & 70\textdegree & 80\textdegree & 90\textdegree       \\
    \midrule
    FR~\cite{ren2016faster} & 94.3 & 90.2  & 86.8 & 80.6  \\
    FR+PIT~\cite{gu2021pit} & 93.6 & 91.4 & 89.2 & 85.9     \\
    
    \textbf{Ours-\emph{h}} & 94.1\tiny{$\pm$ 0.16} & 93.1 \tiny{$\pm$ 0.33} & 91.8 \tiny{$\pm$ 0.40} & 88.8 \tiny{$\pm$ 0.21} \\
    
    \bottomrule
  \end{tabular}
  \caption{FoV Generalization.}
  \label{tab:fovgeneralize}
\end{table}

%% file: tables/mot.tex
% !TEX root = ../main.tex
% !TEX spellcheck = en-US
\begin{figure}
  \centering
  \begin{tabular}{lc}
    \toprule
    Method     & Pedestrian AP@0.5      \\
    \midrule
    FR~\cite{ren2016faster} & 43.7 \\
    AT~\cite{li2022cross} &  63.5\\
    MT  &  64.7 \\
%    Ours-\emph{h} & \\
    \textbf{Ours}     &  \textbf{65.3}\tiny{$\pm$ 0.37} \\  
    \bottomrule
  \end{tabular}
  \captionof{table}{Viewpoint Adaptation.}
 \label{tab:city2mot}
\end{figure}%
\begin{figure}
  \centering
  \includegraphics[width=0.8\linewidth]{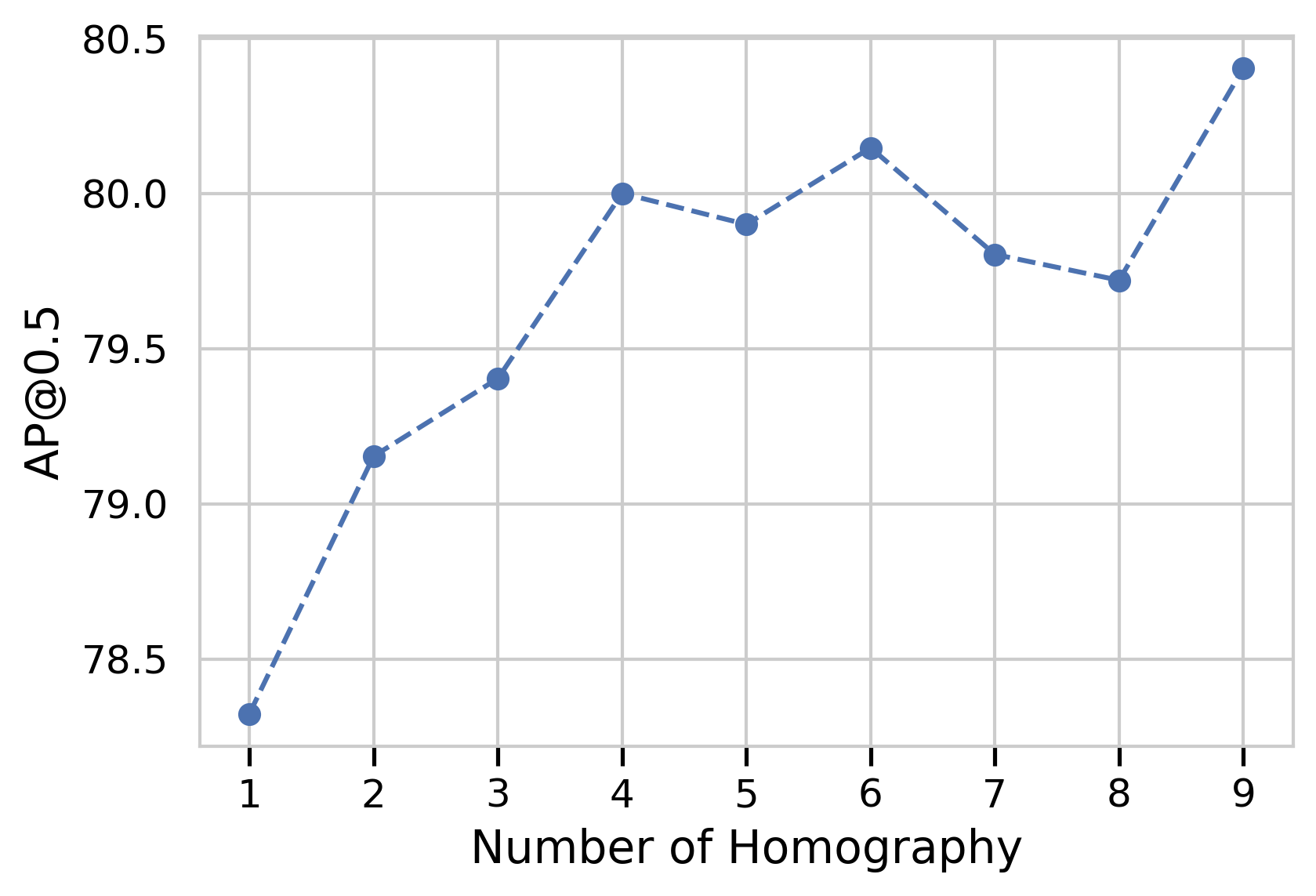}
  \captionof{figure}{\textbf{Varying the number of homographies.} We evaluate the effect of $N$ on the FoV adaptation task. %Each experiment was run three times to produce the box plot above. 
  }
  \label{fig:nstn}

\end{figure}

%% file: figures/featact_small.tex
\begin{figure}[h]
  \centering
  \includegraphics[width=\linewidth]{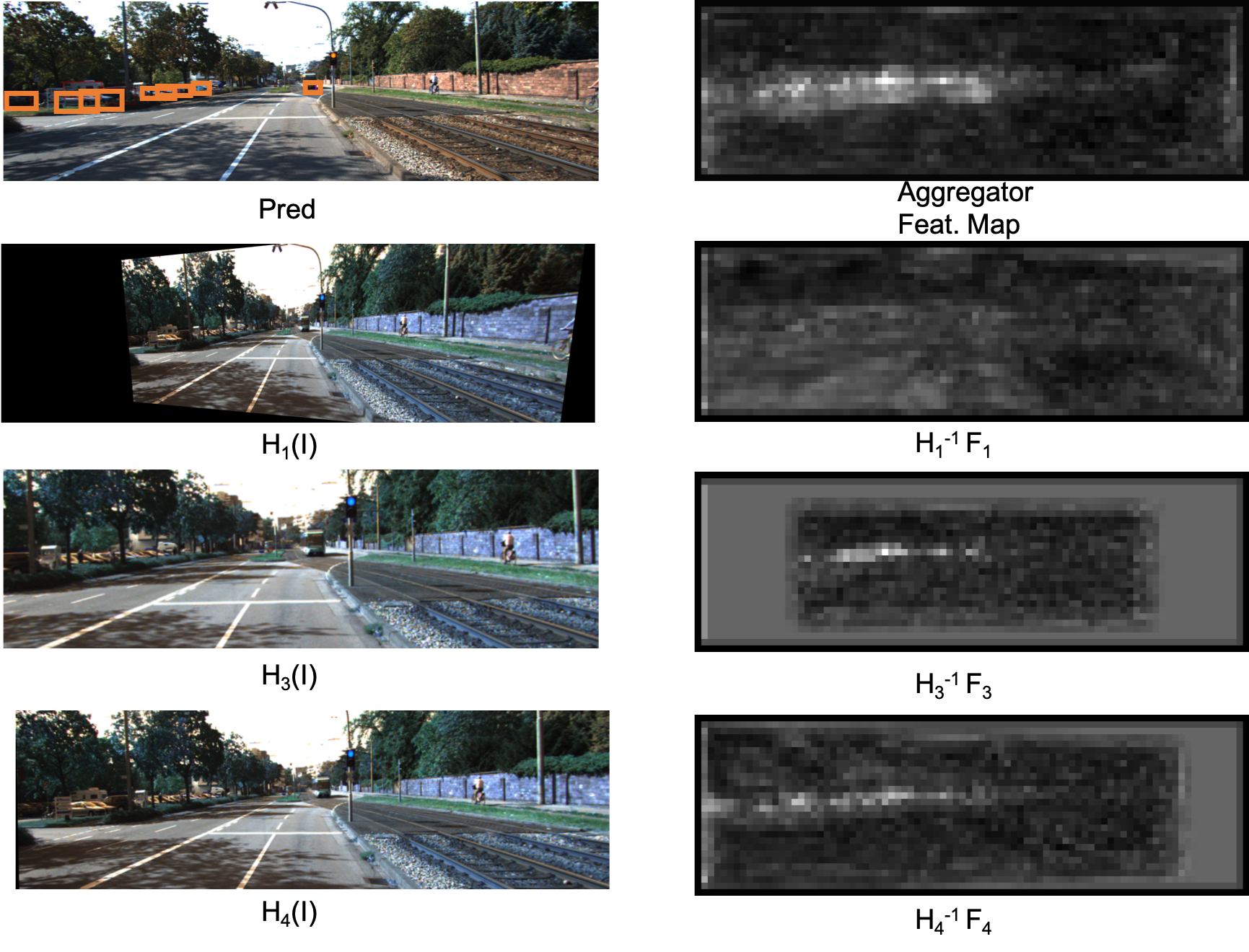}
  \caption{\textbf{Feature Maps}: Top row: predictions of our network and feature map after aggregator. Left column: Image I, transformed by learned homographies; Right Column: Feature maps $F$ warped by corresponding $H^{-1}$ which are input to the aggregator. Each transform distorts the image regions differently. Most of the \emph{cars} are on the left side and of small size in the image. $H_1$ distorts the left side leading to no activation($H_1^{-1}F_1$) for the object. $H_3$ which causes the zoom-in effect has the strongest activation as the smaller objects are visible better here. These maps are generated by taking maximum over channel dimension. } 
  \label{fig:tfactivations_small}
\end{figure}

%% file: figures/nstn.tex
% !TEX root = ../main.tex
% !TEX spellcheck = en-US
% \begin{figure}
%   \centering
%   \includegraphics[width=0.5\textwidth]{figures/nstn.png}
%   \caption{\textbf{Varying the number of homographies.} We evaluate the effect of the number of homographies on the Cityscapes-to-KITTI FoV adaptation task. Each experiment was run three times to produce the box plot above. 
%   }
%   \label{fig:nstn}
% \end{figure}

%% file: figures/evol_from_identity_main.tex
% !TEX root = ../main.tex
% !TEX spellcheck = en-US
\begin{figure}[h]
  \centering
  \includegraphics[width=0.6\linewidth]{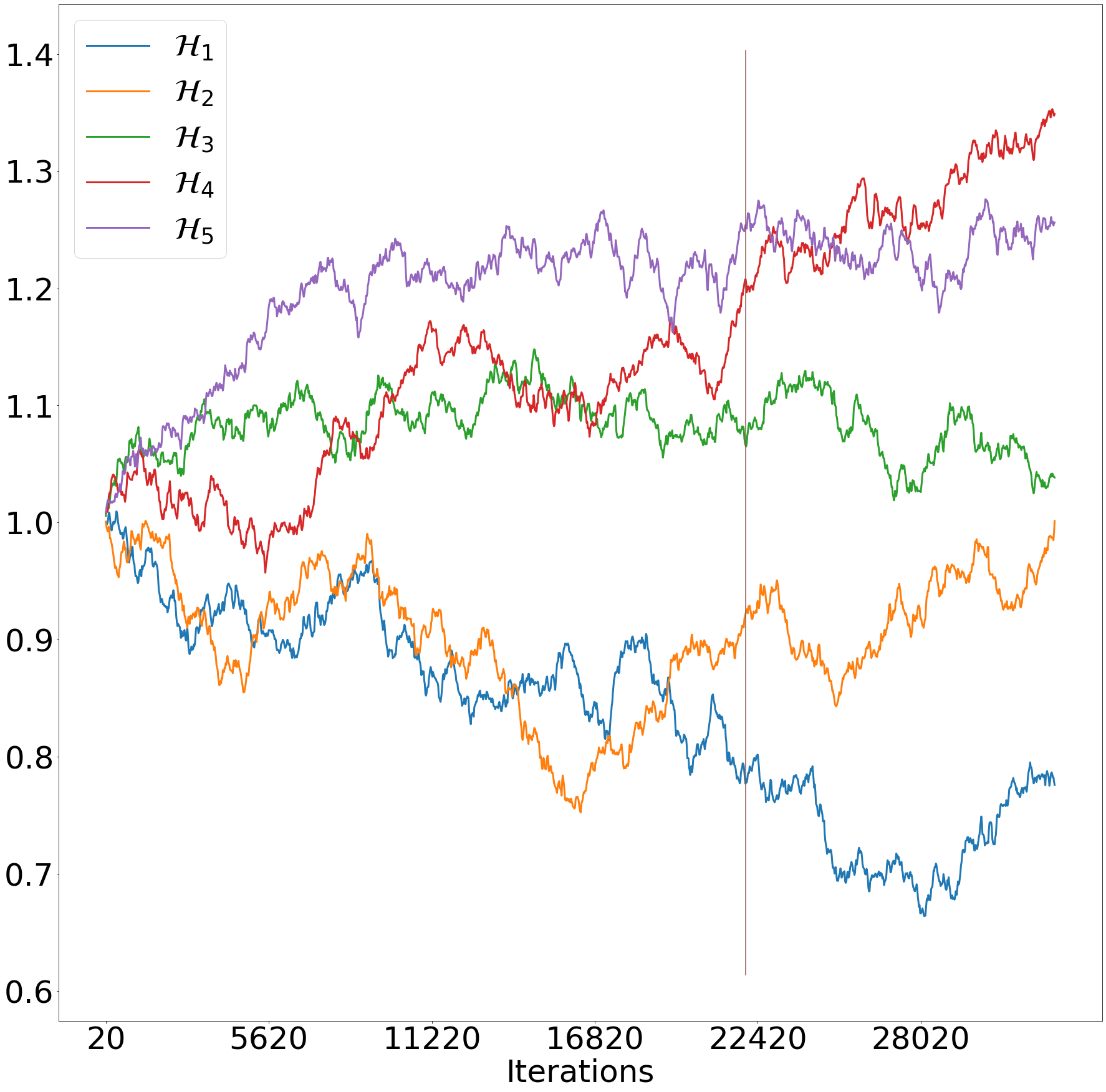}

\caption{\textbf{Diversity in $\mathcal{T}$:} We train $5$ homographies initialized as $\mathcal{H}_i=I$. We plot the evolution of $s_x$ for different homograhies as training proceeds. Each homography is shown in a different color. Note that the values for the different homographies become diverse.  The best score is achieved at iteration = 22k, indicated with the vertical line.}
\label{fig:eval_id_kitti_main}
\end{figure}

%% file: tex/conclusion.tex
\section{Conclusion}
We have introduced an approach to bridge the gap between two domains caused by geometric shifts by learning a set of homographies. We have shown the effectiveness our method on two different kinds of shifts, without relying on any annotations in the target domain, including information about the nature of the geometric shifts. Our analyses have evidenced that optimizing the transformations alone brings in improvement over the base detector and increasing the number of learnt homographies helps further. 
%Our method outperforms the baselines with similar architecture validating our assumption. 
In the future, we plan to learn transformations that are conditioned on the input image to model image-dependent geometric shifts.

%% file: tex/supp.tex
\section{Transformations through Homography}
\input{figures/transformations_example.tex}
We use homography to introduce varied perspective transformations so that they can distort the same image regions differently as seen in~\cref{fig:perspectivtf}. This helps the detector to learn robust object features and simultaneously optimize an aggregator with a different set of homographies which can bridge the gap between two domains.

\section{Feature Maps Activation}
\input{figures/feature_activations.tex}
We show in~\cref{fig:tfactivations} how different homographies generate activation in the feature maps. Not all homographies look at the same image region, therefore the task of the aggregator is to bring in the activations from different transformations together. 

\input{figures/pit_approx_large}

% \newpage
\section{Other Aggregator Architecture}
We implement aggregator using standard functions to combine $\{\mathcal{F}_{\mathcal{H}_i}\}_{i=1}^N$. \cref{tab:supp_diff_aggregator} illustrates this study for FoV adaptation, where the training is done under mean teacher formalism to learn $|\mathcal{T}|=N=5$. We see that these non-learnable aggregators are able to outperform MT baseline (\cref{sec:exp_sota}, in the main paper) suggesting that including transformations helps to bridge the geometric shifts. 
\input{tables/supp_agg_arch} 

% \section{Additional Baseline}

% In Table~\ref{tab:supp_da}, we compare our method with ~\cite{li2022cross}, which also uses mean teacher to adapt appearance based shifts.  We outperform their best model with a big margin, hence showing that the geometric shifts are not bridged  with just feature alignment. We run their publicly available code\footnote{\url{https://github.com/facebookresearch/adaptive_teacher}} for ResNet50 backbone. 
% \input{tables/supp_da}

% \section{Diversity in $\mathcal{T}$}
\section{Diversity in T}
\label{sec:diversityinT}
In order to show that diverse transformations are learned, we set $\mathcal{H}_i$ = $I$ and train our mean teacher formulation.  \cref{fig:eval_id_kitti} shows diverse set of transformations learned in FoV adaptation task. Even though we do not enforce diversity among homographies, it is learned through our approach.

\input{figures/evol_from_identity}

% \section{Evolution of $\mathcal{T}$}
\section{Evolution of T}
\label{sec:evolutionofT}
We provide qualitative results for $\mathcal{T}$  learned in FoV and Viewpoint adaptation, \cref{fig:eval_quant_kitti} and \cref{fig:eval_quant_mot}, respectively. The qualitative results for the same adaptation task can be seen in \cref{fig:eval_kitti} and \cref{fig:eval_mot}, respectively. 
\input{figures/evol_kitti}
\input{figures/eval_kitti}

\input{figures/eval_mot}
\input{figures/evol_mot}

\input{figures/evolution.tex}
\input{figures/sota_mot_visu}
\section{Hyperparameter details}
\paragraph{Augmentations.} We use Detectron2\cite{wu2019detectron2}s implementation for random crop and torchvision~\footnote{\url{https://pytorch.org/vision/stable/transforms.html}} for color jittering.
\input{tables/supp_aug}

\paragraph{FasterRCNN~\cite{ren2016faster} training.} We train our base network with random crop strategy on with only source data, which is Cityscapes for both the adaptation tasks. The trained model achieves 74.7 and 58.4 AP@0.5 score on the source domain validation set for \emph{car} and \emph{person} detection, respectively.

\paragraph{Mean Teacher Training} For our mean teacher setup (\cref{sec:method_training}, in the main paper), we choose $\tau=0.6$ as the confidence threshold for the pseudo-labels and evaluate contribution of target domain loss for different $\lambda$. \cref{fig:supp_lambda} summarizes this study. We see that method performs worse when we have equal contribution from both source and target domain loss $\lambda=1$, as the false positives in the target domain quickly deteriorate the training. \cref{fig:supp_tau}, evaluation for different values of $\tau$.
\input{figures/lamda_supp}
\input{figures/tau_supp}
\section{Architecture details}
Our aggregator architecture consists of three convolution layers along with BatchNorm and Relu layers after each convolution. \cref{tab:supp_aggregator} shows the details of different layers. Here, $C=1024$ corresponds to the output of the feature extractor.
\input{tables/supp_arch_details}

% \section{Assets and licenses}
% Every asset used is publicly available under different licenses. In this section we list the different assets and put a reference to their respective license.
% \begin{itemize}
%     \item Kitti dataset \cite{geiger2012we} - CC BY-NC-SA 3.0 \footnote{http://www.cvlibs.net/datasets/kitti/}
%     \item MOT20 dataset \cite{dendorfer2020mot20} - CC BY-NC-SA 3.0 \footnote{https://motchallenge.net/}
%     \item Cityscape dataset \cite{cordts2016cityscapes} - Custom license \footnote{https://www.cityscapes-dataset.com/license/}
%     \item Detectron2 \cite{wu2019detectron2} code library - Apache 2.0 \footnote{https://github.com/facebookresearch/detectron2/blob/main/LICENSE}
% \end{itemize}

%% file: figures/transformations_example.tex
\begin{figure*}[h]
  \centering
  \includegraphics[width=\textwidth]{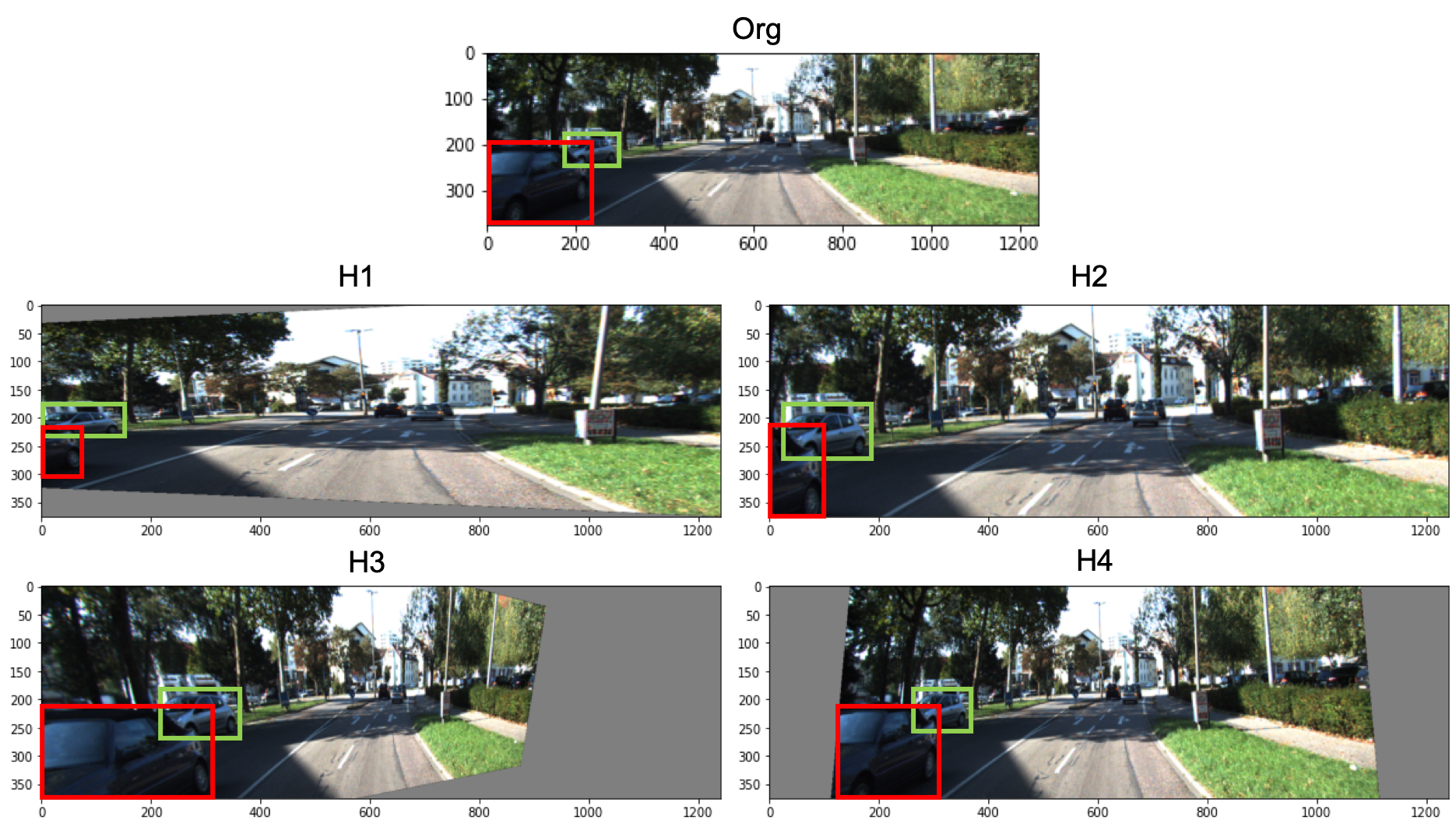}
  \caption{\textbf{Transformations}: Here we demonstrate how the two objects in the original image undergo different perspective transformations. Our task is to learn robust object features under such transformations and use them to bring the two domains closer while being agnostic to the camera parameters. We train with a multiple set of transformations to change the same image region differently. With our trainable aggregator, we can then combine features from different regions to help in improving the detector's performance.}
  \label{fig:perspectivtf}
\end{figure*}

%% file: figures/feature_activations.tex
\begin{figure*}[h]
  \centering
  \includegraphics[width=\textwidth]{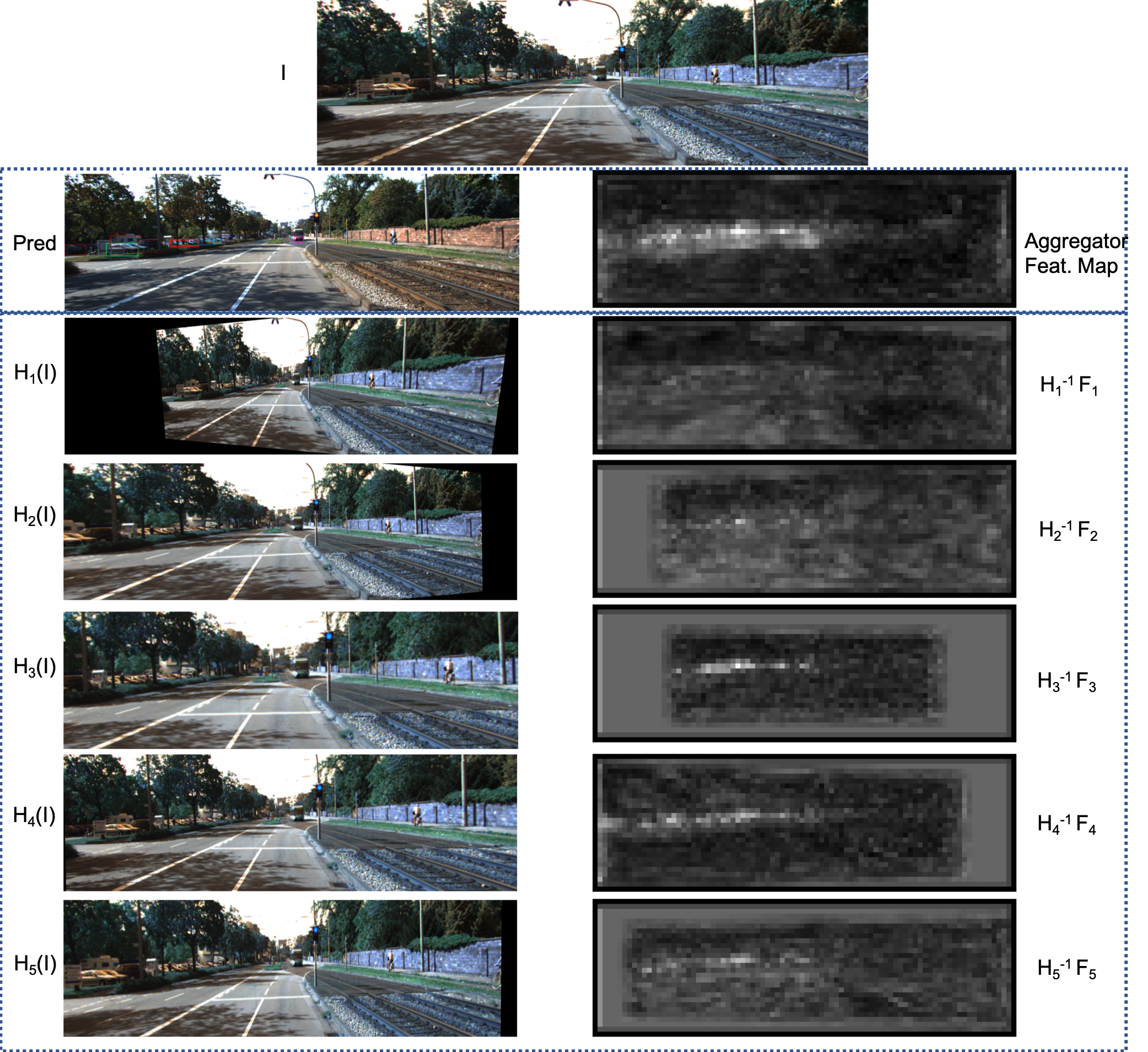}
  \caption{\textbf{Feature Maps}: Top row: predictions of our network and feature map after aggregator. Left column: Image I, transformed by 5 learnt homographies; Right Column: Feature maps $F$ warped by corresponding $H^{-1}$ which are input to aggregator. Each transform distorts the image regions differently. Most of the \emph{cars} are on the left side and of small size in the image. $H_1$ distorts the left side leading to no activation($H_1^{-1}F_1$) for the object. $H_3$ which causes zoom-in effect has the strongest activation as the smaller objects are visible better here. Overall aggregator feature map contains activation from the region where the objects exist. The aggregator has learnt how to combine regions with activations under different homographies. The feature maps are generated by taking maximum over channel dimension.} 
  \label{fig:tfactivations}
\end{figure*}

%% file: figures/pit_approx_large.tex
\begin{figure*}[h]
  \centering
  \includegraphics[width=\textwidth]{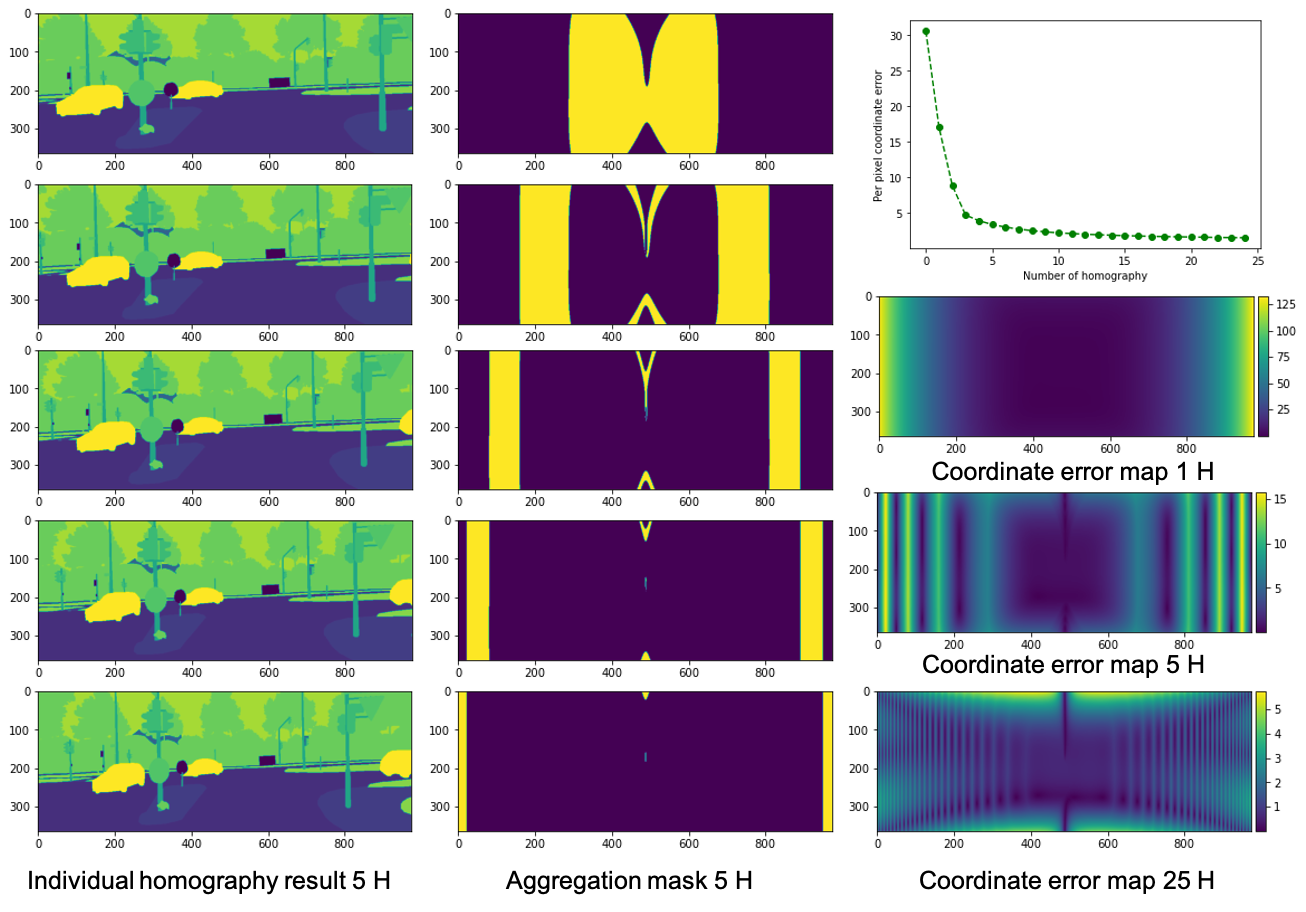}
  \caption{{\bf Approximating PIT with homographies.} Left column: Visualization of each homography use to approximate PIT with 5 transforms; the top one is the identitity, and the following ones are in order of increasing compression. Center column: Contribution of each homography to the final remapping. Right column: The top figure shows the per pixel coordinate error when compared to the PIT remapping as a function of the number of homographies used in the approximation; the three bottom figures depict the coordinate error maps for 1, 5, and 25 homographies used to approximate PIT (note the scale change in pixel coordinate error).}
  \label{fig:pit_approx}
\end{figure*}

%% file: tables/supp_agg_arch.tex
% !TEX root = ../supp.tex
% !TEX spellcheck = en-US
\begin{table}[h]
  \centering
  \begin{tabular}{lcc}
    
    Function     & Car AP@0.5 \\
    \midrule
    sum &  78.1\tiny{$\pm$ 0.14} \\
    mean & 78.7\tiny{$\pm$ 0.05} \\
    max & 78.7\tiny{$\pm$ 0.12} \\
    min+max & 78.9\tiny{$\pm$ 0.43} \\
    \bottomrule
    MT & 78.3 \\
    Ours & 79.9\tiny{$\pm$ 0.14} \\
    \bottomrule
  \end{tabular}
    \caption{Aggregator Architecture without learnable parameters}
  \label{tab:supp_diff_aggregator}
\end{table}

%% file: figures/evol_from_identity.tex
% !TEX root = ../supp.tex
% !TEX spellcheck = en-US
\begin{figure*}[h]
\begin{minipage}{0.5\textwidth}
  \centering
    \captionsetup{labelformat=empty}
  \caption{$s_x$}
  \addtocounter{figure}{-1}
  \includegraphics[width=\textwidth]{figures/id_sx.png}

  \label{fig:id_sx}
\end{minipage}%
\begin{minipage}{0.5\textwidth}
    \centering
    \captionsetup{labelformat=empty}
    \caption{ $s_y$}
    \addtocounter{figure}{-1}
  \includegraphics[width=\textwidth]{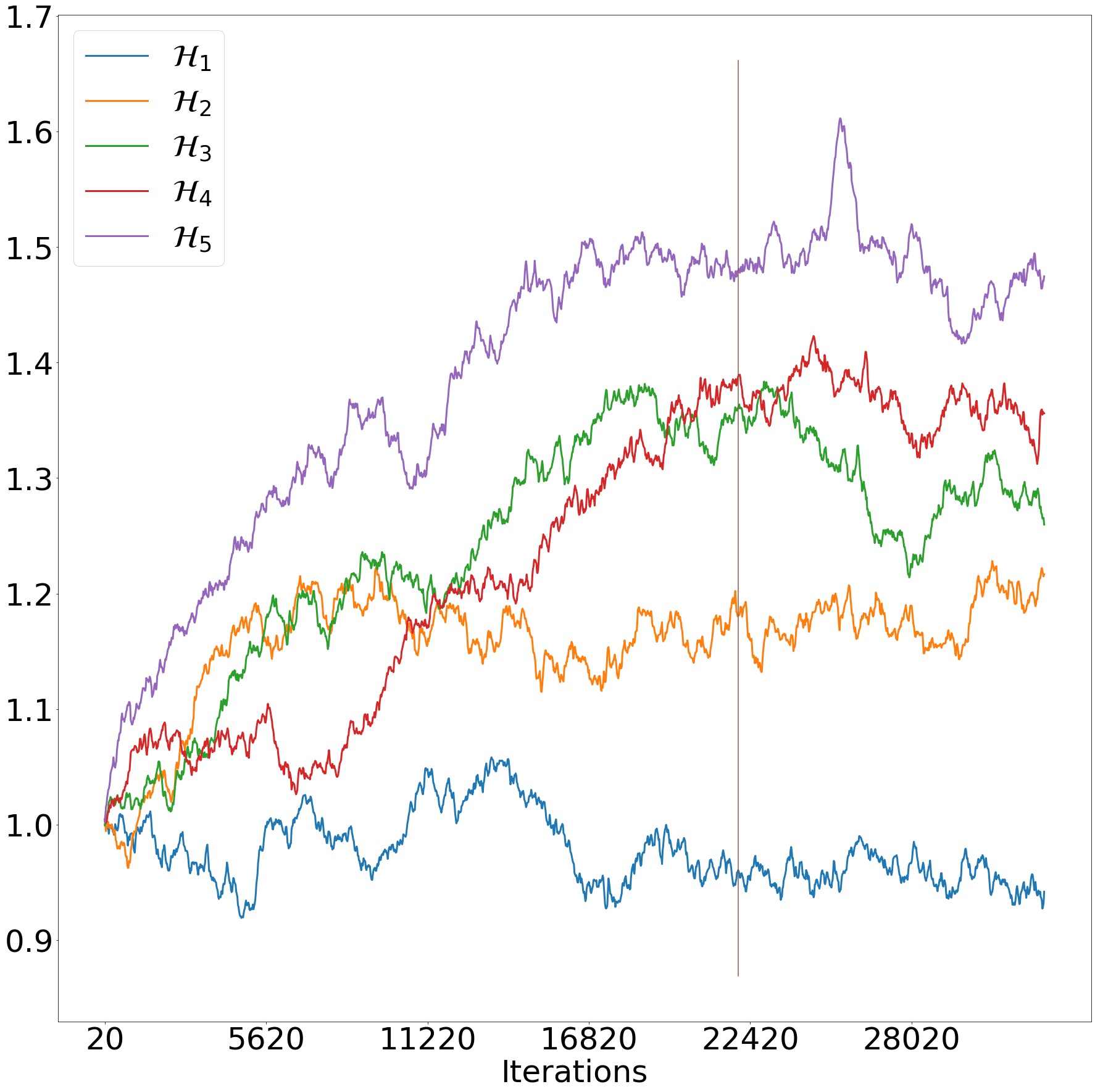}

  \label{fig:id_sy}
 \end{minipage}
  \vspace{1cm}
  
 \begin{minipage}{0.5\textwidth}
    \centering
   \captionsetup{labelformat=empty}
  \caption{ $l_x$}
  \addtocounter{figure}{-1}
  \includegraphics[width=\textwidth]{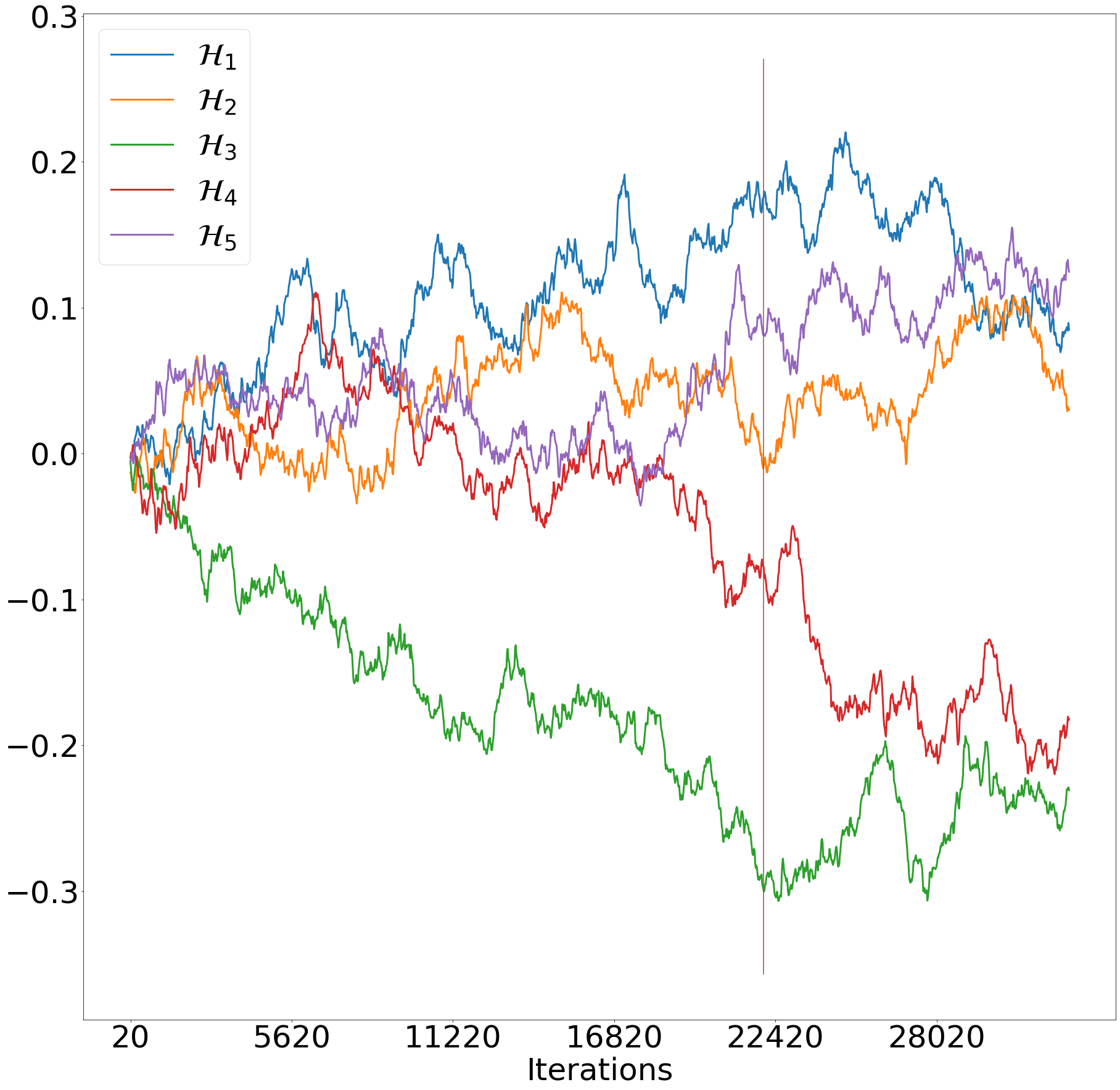}

  \label{fig:id_lx}
 \end{minipage}%
 \begin{minipage}{0.5\textwidth}
    \centering
      \captionsetup{labelformat=empty}
  \caption{ $l_y$}
  \addtocounter{figure}{-1}
  \includegraphics[width=\textwidth]{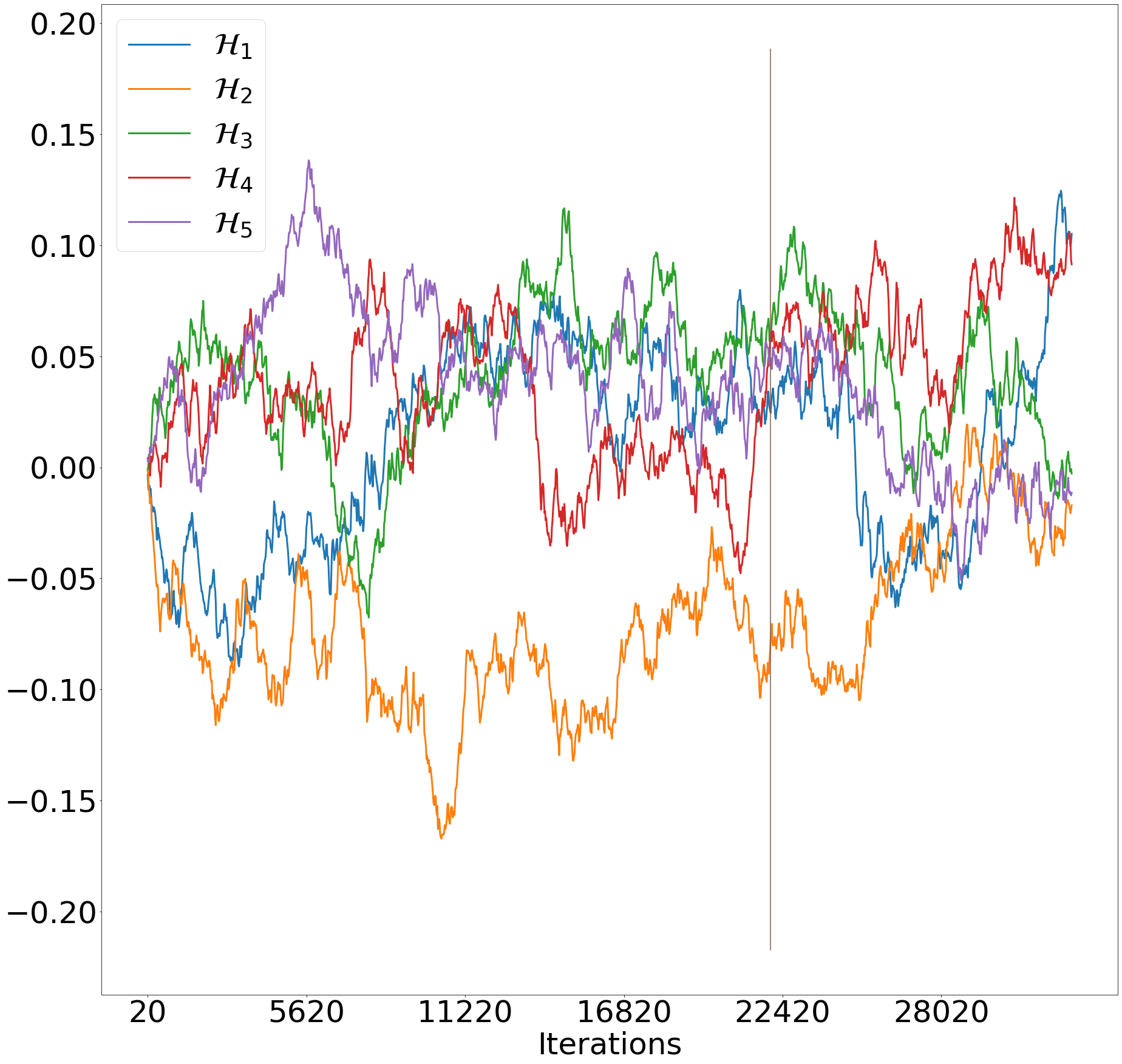}

  \label{fig:id_ly}
 \end{minipage}
\caption{\textbf{Diversity in $\mathcal{T}$:} We train $|\mathcal{T}|=5$ initialized with $\mathcal{H}_i=I$. Homographies parameterized by $s_x,s_y,l_x,l_y$ evolve as the training proceeds and tend to become diverse. Each homography is shown in different color. Even though we do not enforce any diversity, our approach learns diverse set of transformations. With these learned homorgraphies, we achieve 79.5 AP@0.5 score for FoV adaptation task. The best score is achieved at iteration = 22k shown with the vertical line.}
\label{fig:eval_id_kitti}
\end{figure*}

%% file: figures/evol_kitti.tex
% !TEX root = ../supp.tex
% !TEX spellcheck = en-US
\begin{figure*}[h]
\begin{minipage}{0.5\textwidth}
  \centering
  \captionsetup{labelformat=empty}
  \caption{$s_x$}
  \addtocounter{figure}{-1}
  \includegraphics[width=\textwidth]{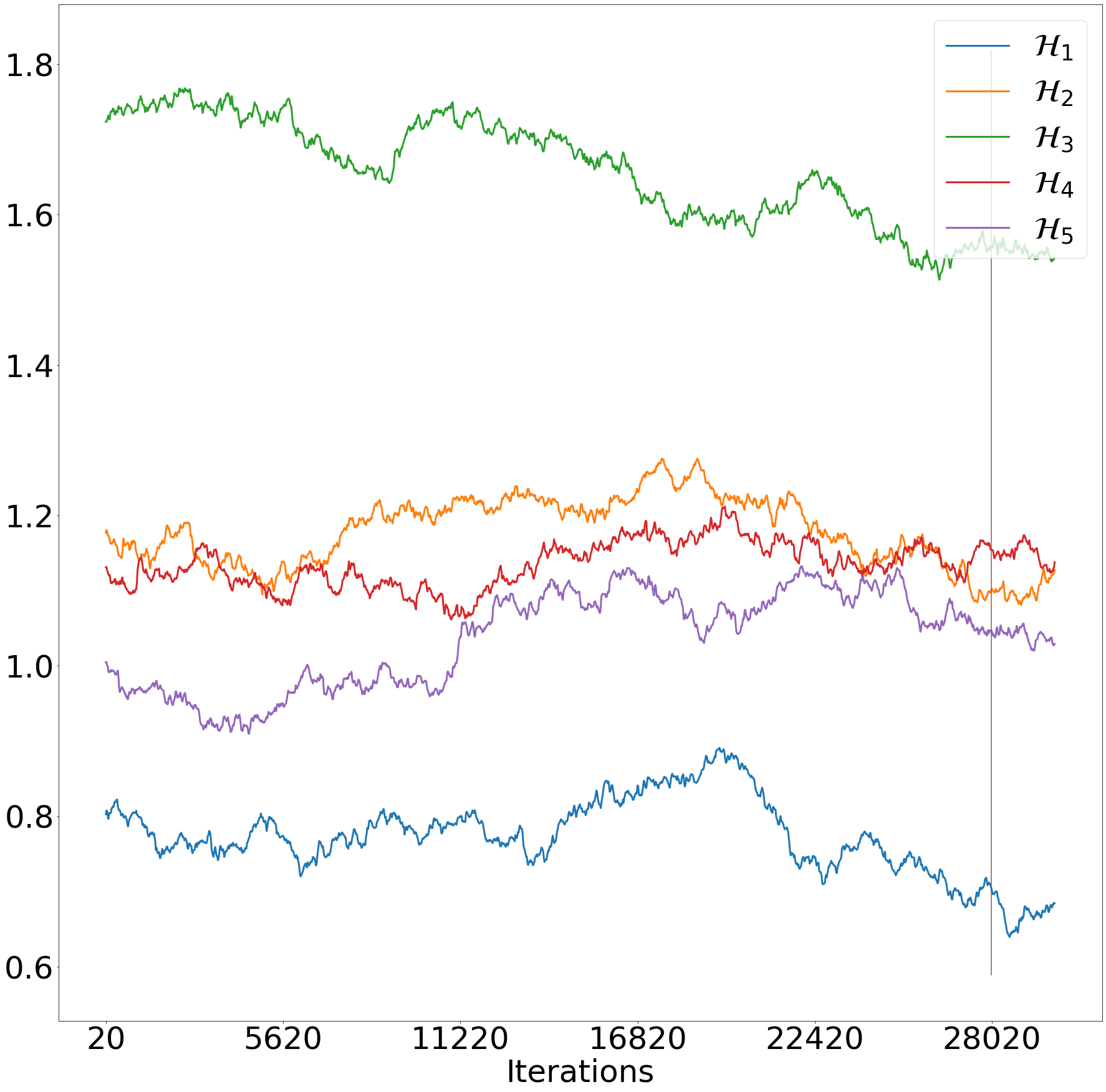}

  \label{fig:kitti_sx}
\end{minipage}%
\begin{minipage}{0.5\textwidth}
    \centering
   \captionsetup{labelformat=empty}
  \caption{ $s_y$}
  \addtocounter{figure}{-1}
  \includegraphics[width=\textwidth]{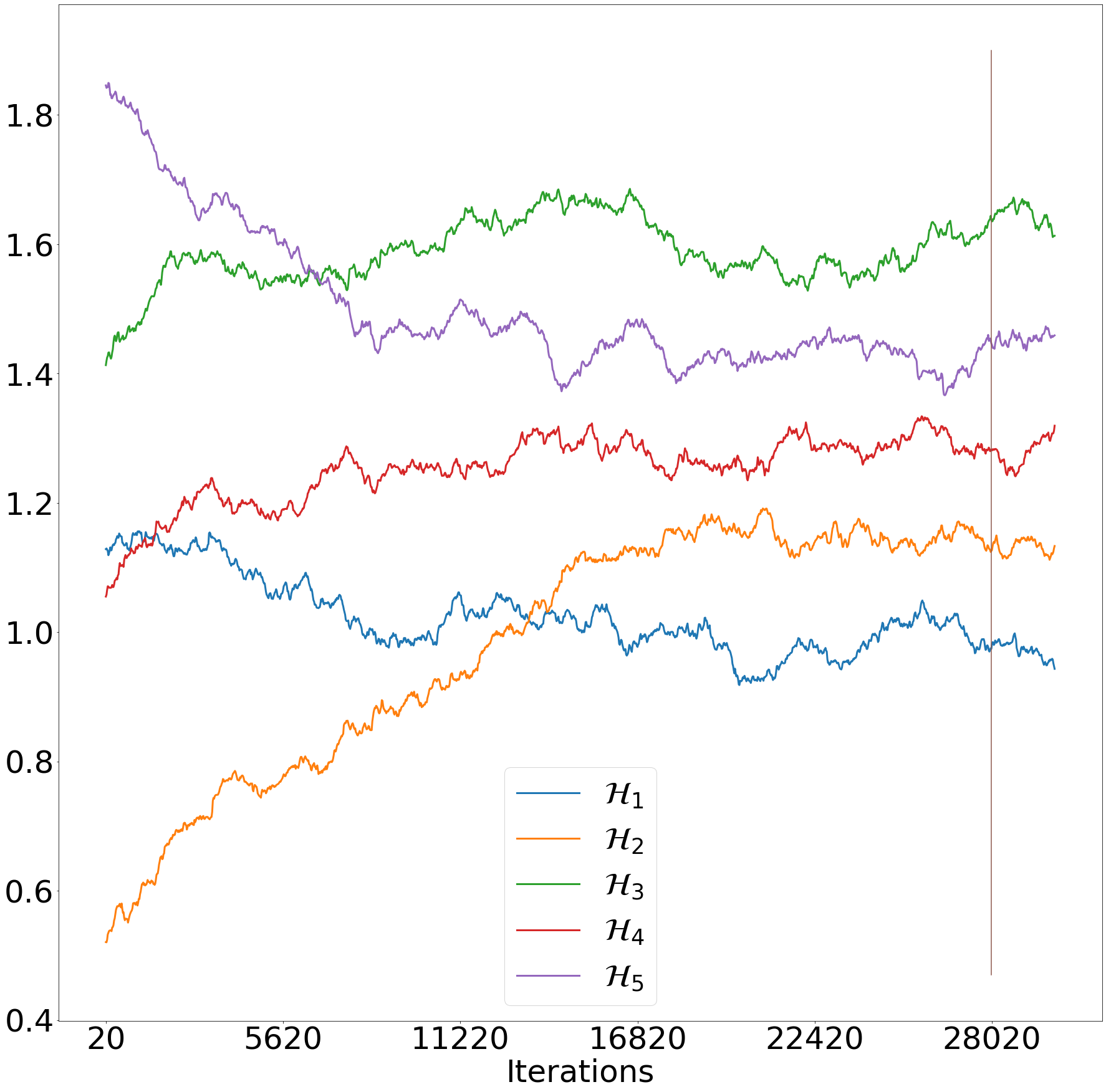}

  \label{fig:kitti_sy}
 \end{minipage}
 \vspace{1cm}

 \begin{minipage}{0.5\textwidth}
    \centering
    \captionsetup{labelformat=empty}
  \caption{ $l_x$}
  \addtocounter{figure}{-1}
  \includegraphics[width=\textwidth]{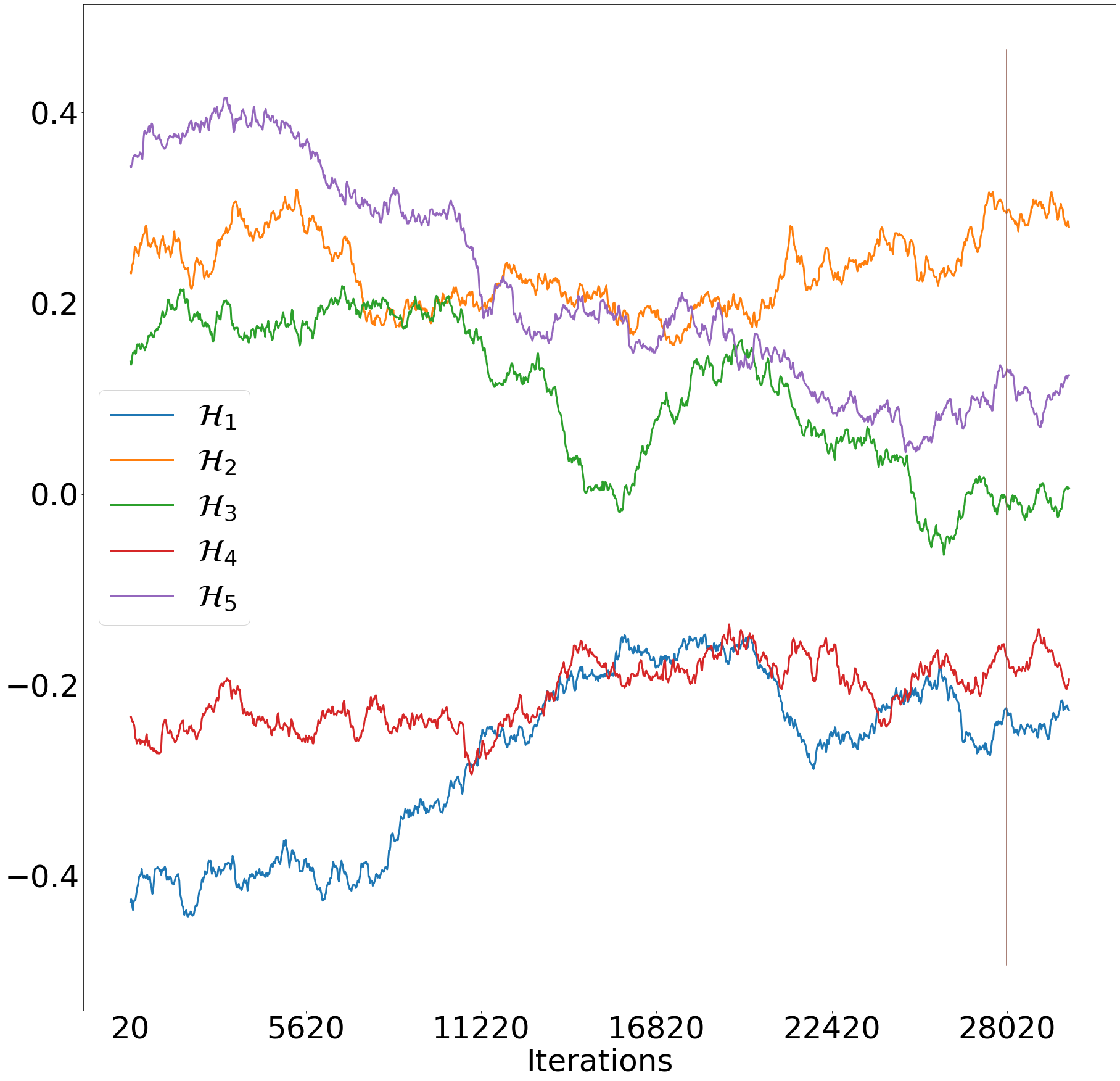}

  \label{fig:kitti_lx}
 \end{minipage}%
 \begin{minipage}{0.5\textwidth}
    \centering
    \captionsetup{labelformat=empty}
  \caption{ $l_y$}
  \addtocounter{figure}{-1}
  \includegraphics[width=\textwidth]{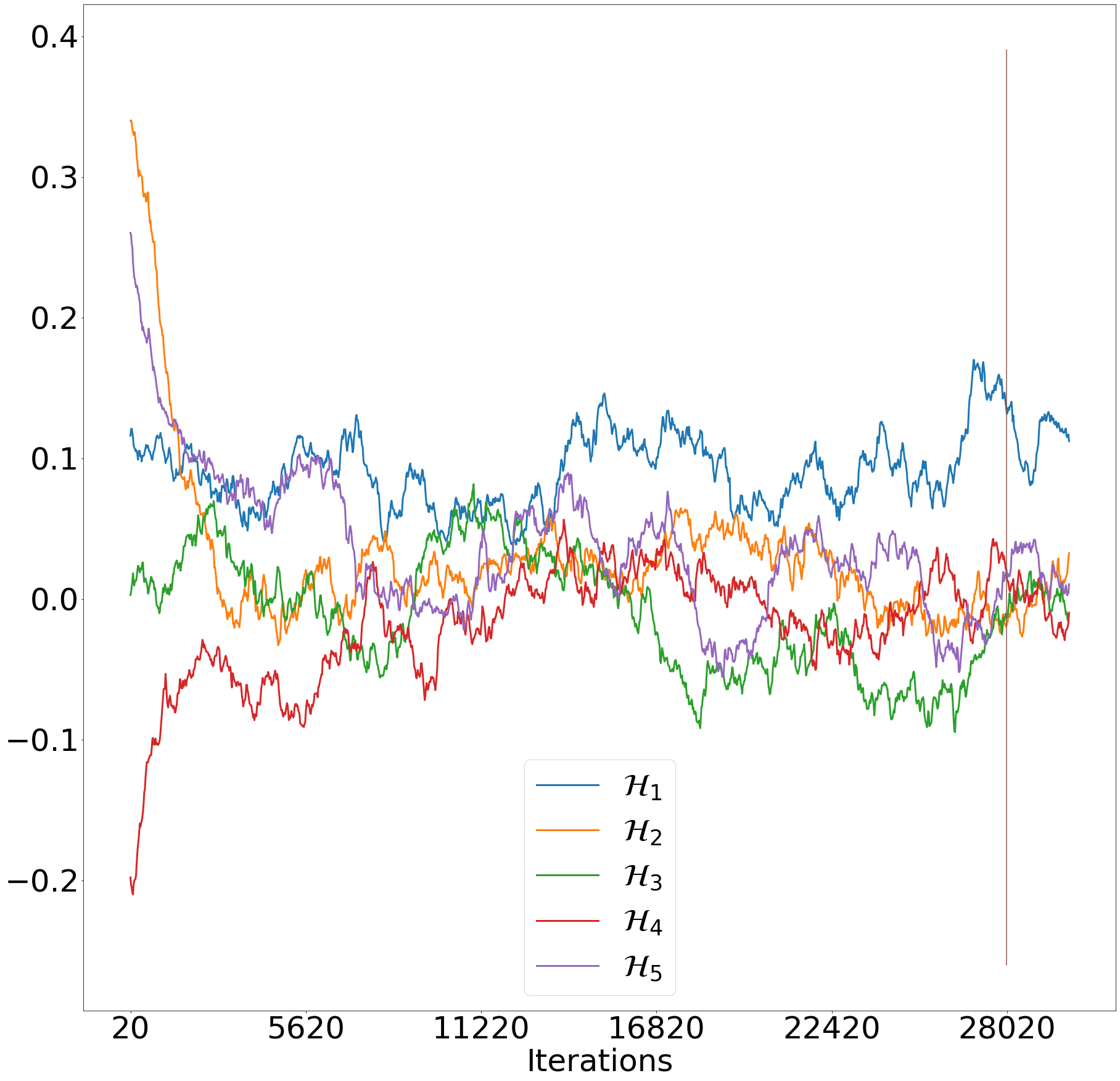}

  \label{fig:kitti_ly}
 \end{minipage}
\caption{Quantitative results for the corresponding results in Figure~\ref{fig:eval_kitti}. The randomly initialized transforms, parameterized by $s_x,s_y,l_x,l_y$, evolve to achieve the best score at 28k iterations (shown by the vertical bar). The colors represent different homographies. Some set of parameters converges to similar value but overall each homography is unique.  }
\label{fig:eval_quant_kitti}
\end{figure*}

%% file: figures/eval_kitti.tex
% !TEX root = ../supp.tex
% !TEX spellcheck = en-US
\begin{figure*}[h]
    \centering
    \includegraphics[width=\textwidth]{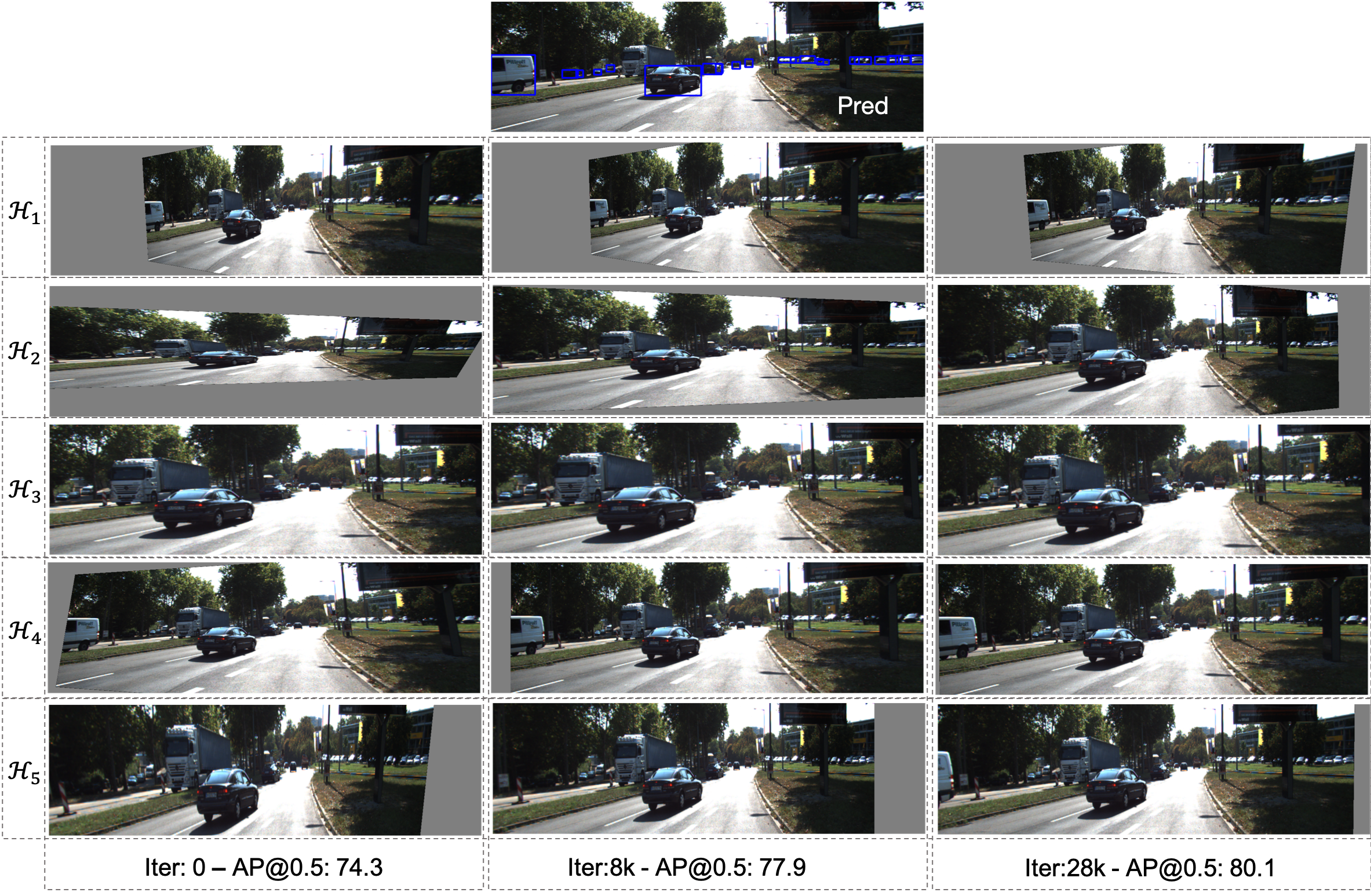}
    \caption{\textbf{FoV adaptation:} The randomly initialized homographies evolve as the training progresses to improve the overall AP score. We train with 5 homographies and show how they transform an image for the corresponding FoV adaptation task. }
    \label{fig:eval_kitti}
\end{figure*}

%% file: figures/eval_mot.tex
% !TEX root = ../supp.tex
% !TEX spellcheck = en-US
\begin{figure*}[h]
    \centering
    \includegraphics[width=\textwidth]{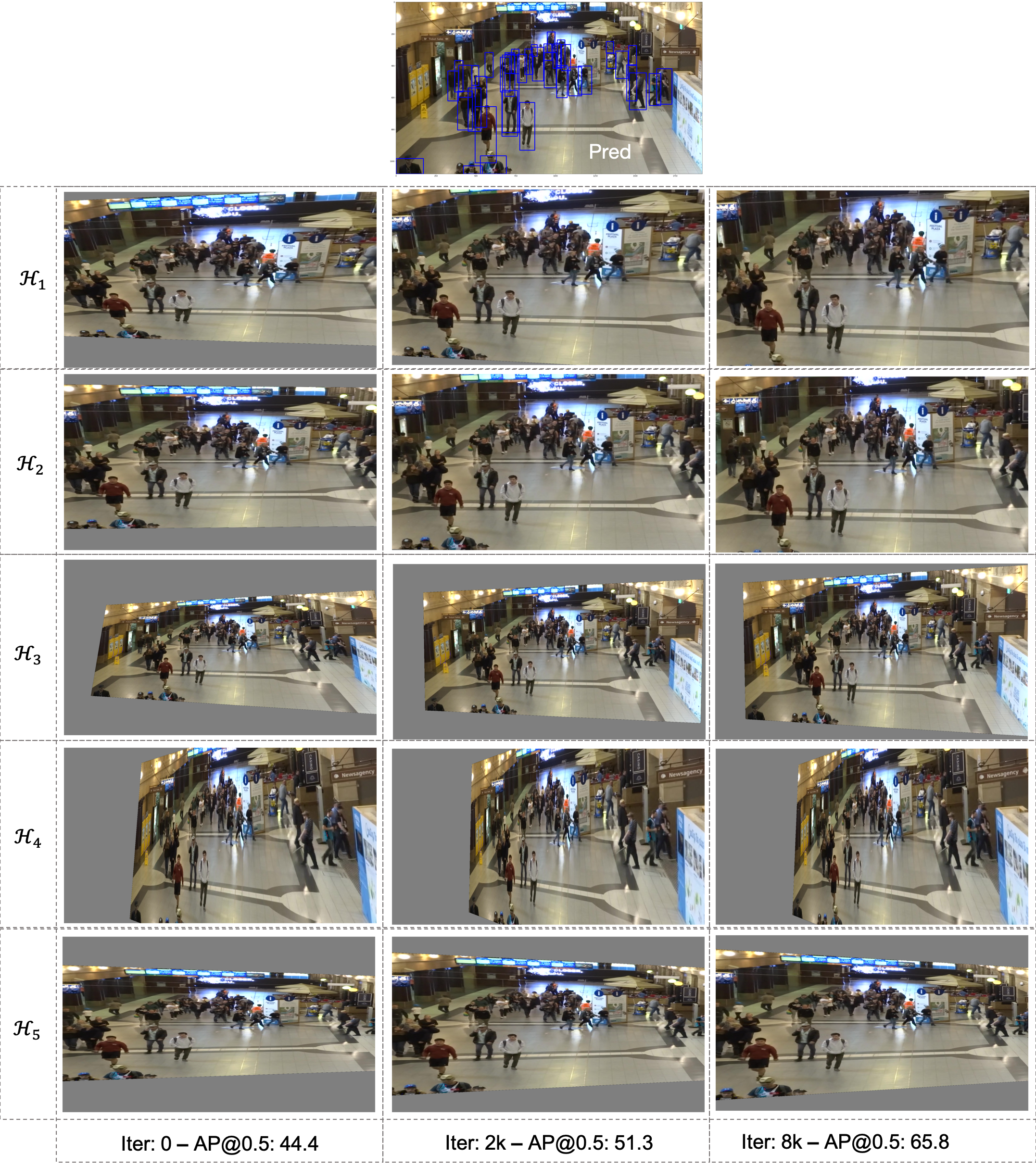}
    \caption{\textbf{Viewpoint adaptation:} The randomly initialized homographies evolve as the training progresses to improve the overall AP score. We train with 5 homographies and show how they transform an image for the corresponding viewpoint adaptation task. }
    \label{fig:eval_mot}
\end{figure*}

%% file: figures/evol_mot.tex
% !TEX root = ../supp.tex
% !TEX spellcheck = en-US
\begin{figure*}[h]
\begin{minipage}{0.5\textwidth}
  \centering
  \captionsetup{labelformat=empty}
  \caption{$s_x$}
  \addtocounter{figure}{-1}
  \includegraphics[width=\textwidth]{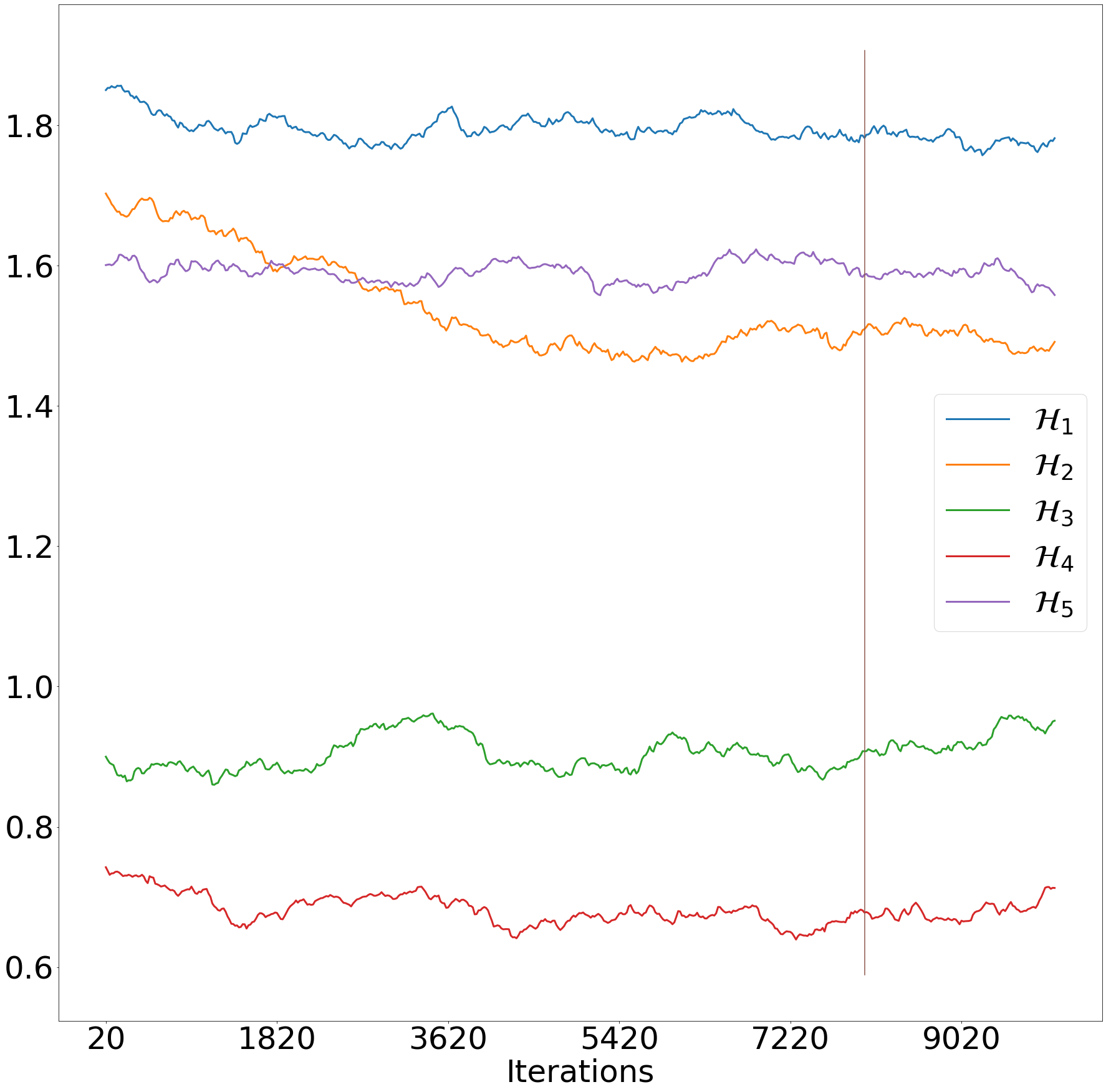}

  \label{fig:mot_sx}
\end{minipage}%
\begin{minipage}{0.5\textwidth}
    \centering
    \captionsetup{labelformat=empty}
  \caption{ $s_y$}
  \addtocounter{figure}{-1}
  \includegraphics[width=\textwidth]{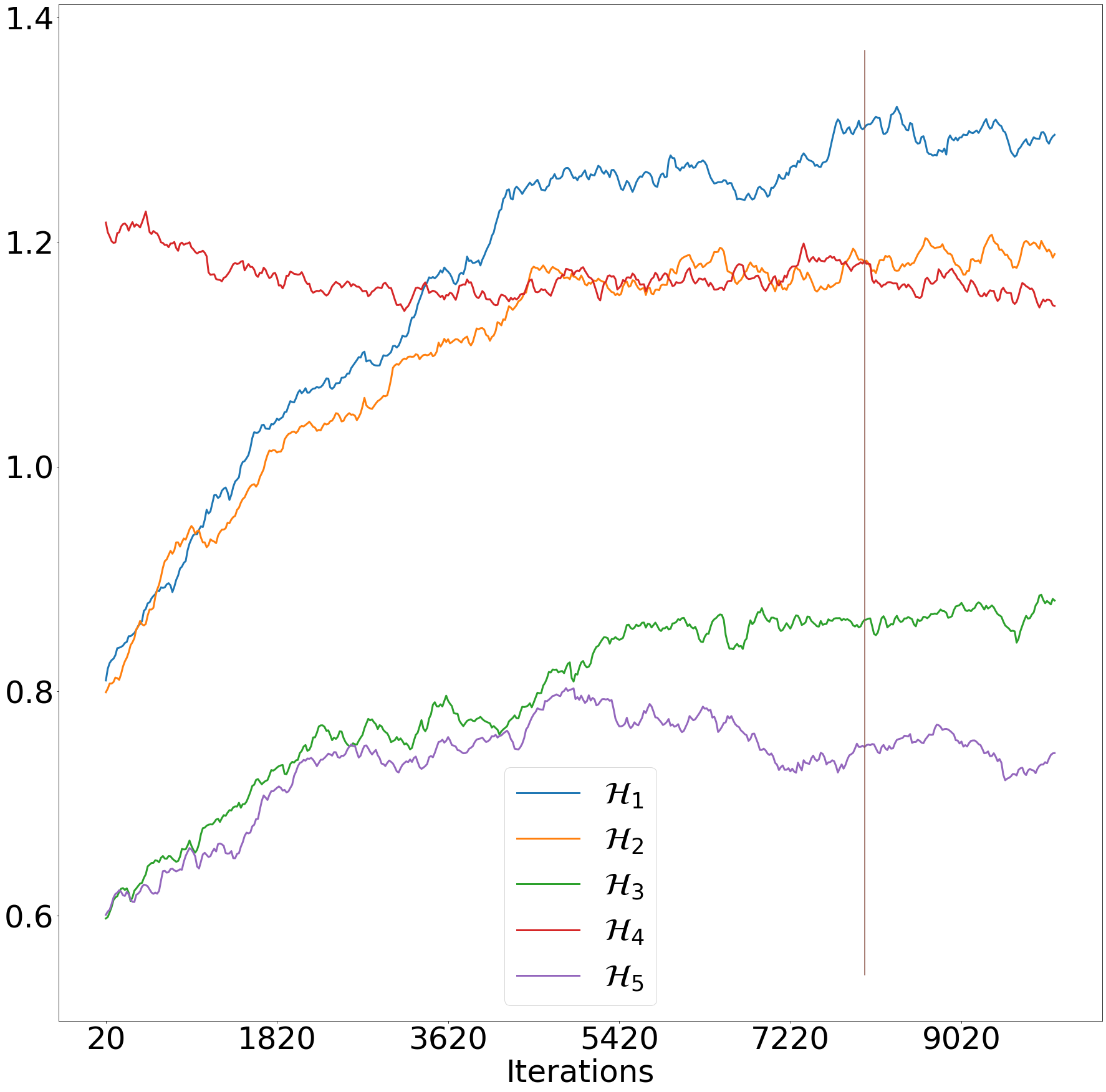}

  \label{fig:mot_sy}
 \end{minipage}
  \vspace{1cm}
  
 \begin{minipage}{0.5\textwidth}
    \centering
    \captionsetup{labelformat=empty}
  \caption{ $l_x$}
  \addtocounter{figure}{-1}
  \includegraphics[width=\textwidth]{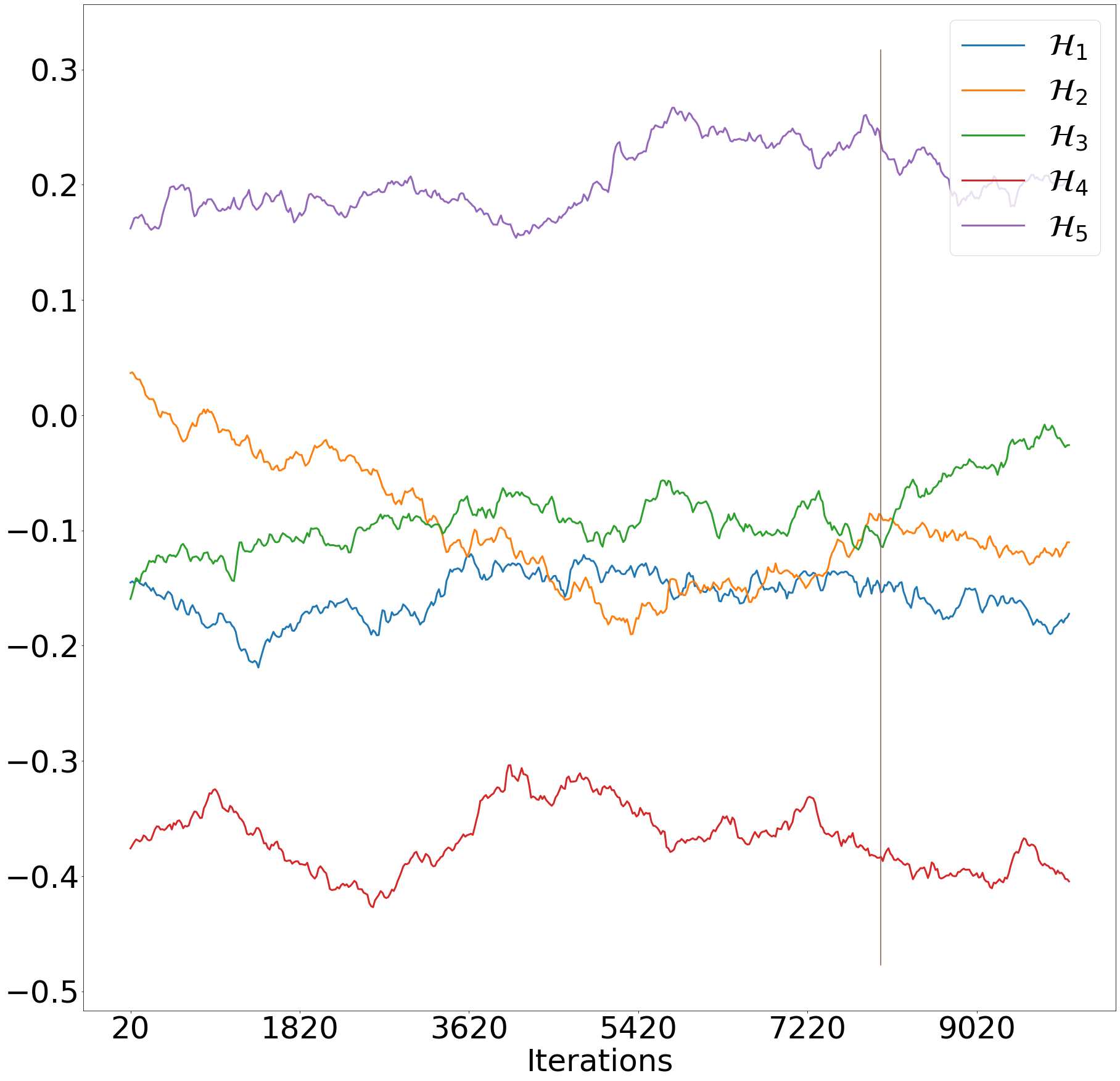}

  \label{fig:mot_lx}
 \end{minipage}%
 \begin{minipage}{0.5\textwidth}
    \centering
      \captionsetup{labelformat=empty}
  \caption{ $l_y$}
  \addtocounter{figure}{-1}
  \includegraphics[width=\textwidth]{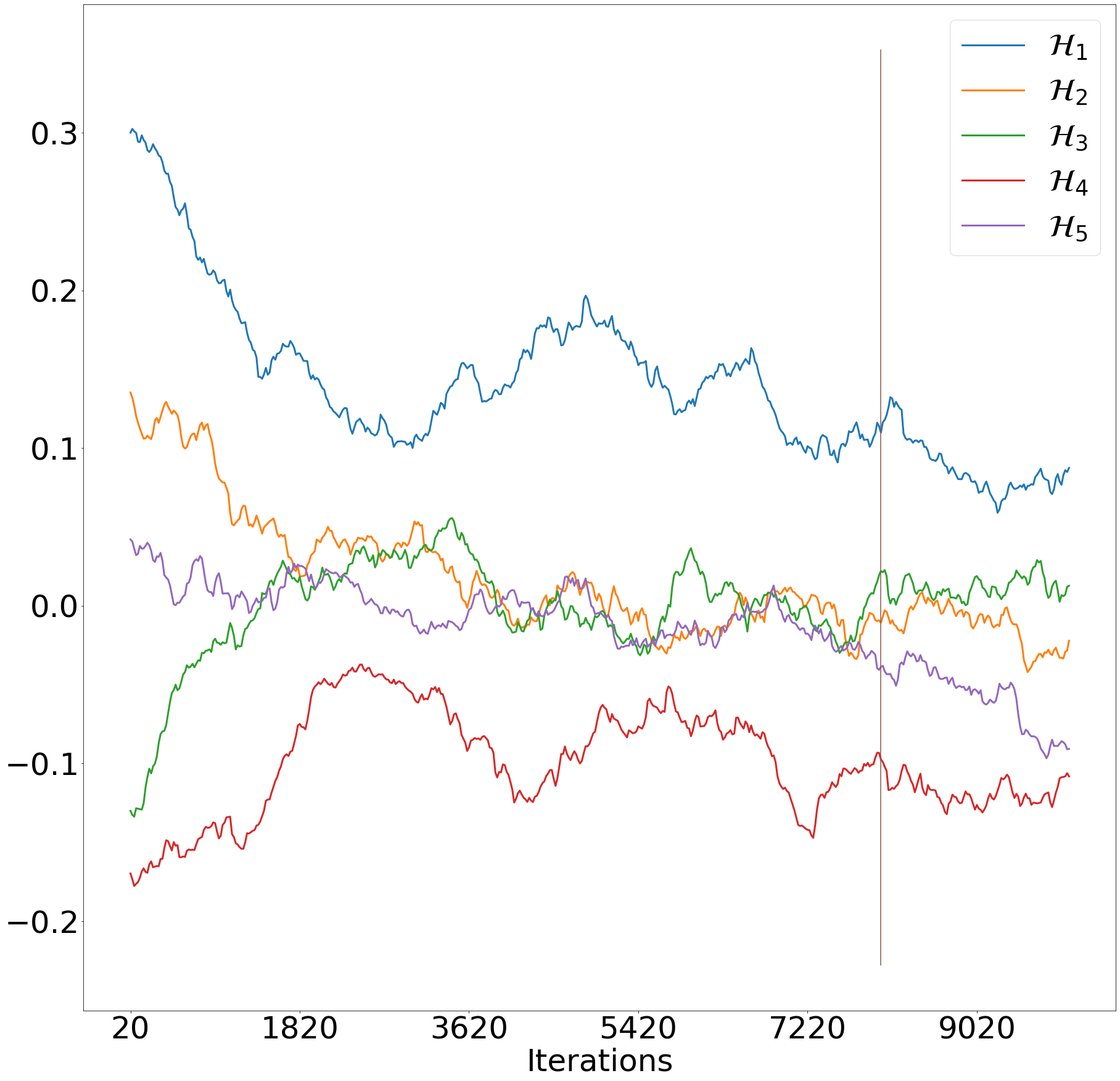}

  \label{fig:mot_ly}
 \end{minipage}
\caption{Quantitative results for the corresponding results in Figure~\ref{fig:eval_mot}. The randomly initialized transforms, parameterized by $s_x,s_y,l_x,l_y$, evolve to achieve the best score at 8k iterations (shown by the vertical bar). The colors represent different homographies. Some $s_y$ parameters start at a similar value but eventually diverge. }
\label{fig:eval_quant_mot}
\end{figure*}

%% file: figures/evolution.tex
% !TEX root = ../main.tex
% !TEX spellcheck = en-US
\begin{figure*}
  \centering
  \includegraphics[width=\textwidth]{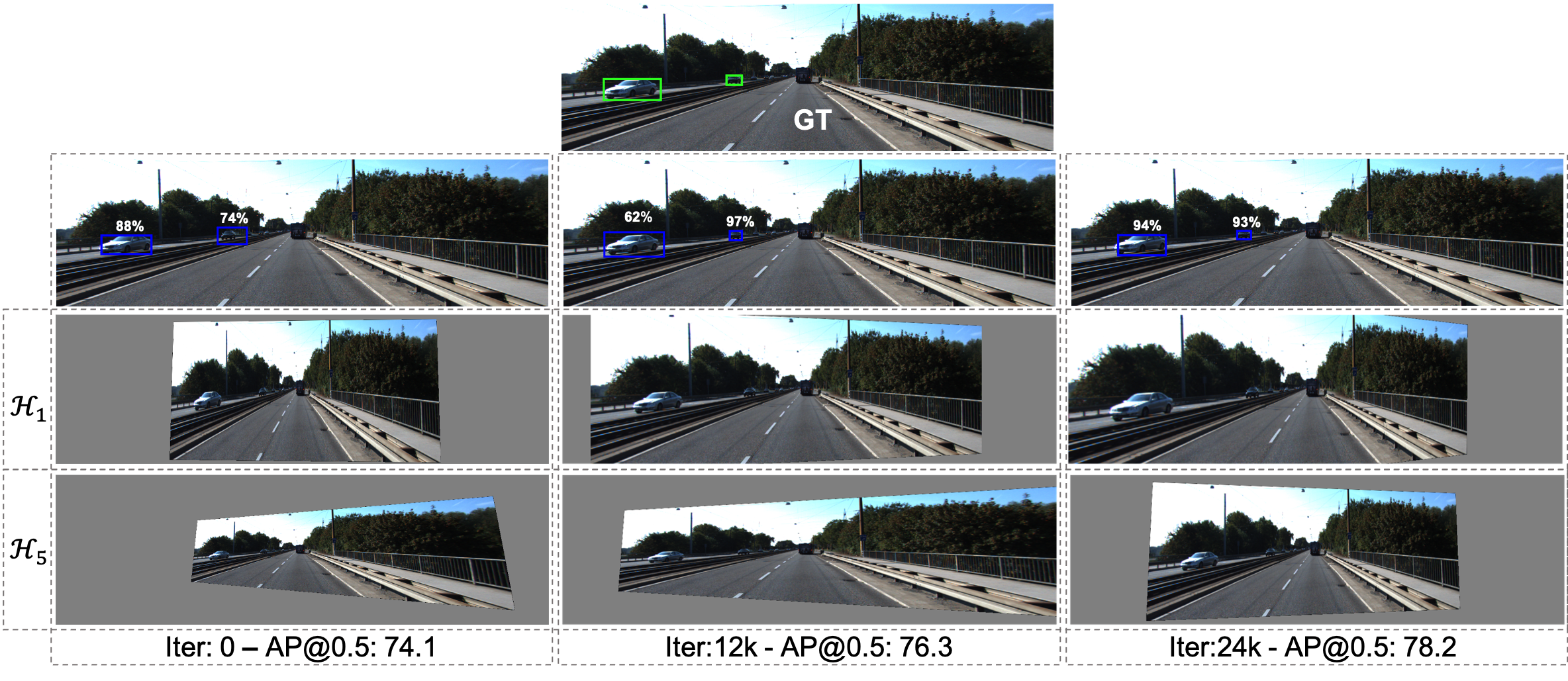}
  \caption{\textbf{Evolution of} $\mathcal{T}$. We showcase how two homographies, $\mathcal{H}_1$ and $\mathcal{H}_5$, evolve across the training iterations and influence the prediction scores. Starting from random homographies at iteration 0, the transformations converge to homographies suited for FoV adaptation. The detection scores consequently increase throughout the training process. Moreover, this increase in detection score is reflected in the overall AP@0.5 score, which jumps from 74.1 to 78.2.}
  \label{fig:evolution}
\end{figure*}

%% file: figures/sota_mot_visu.tex
% !TEX root = ../main.tex
% !TEX spellcheck = en-US
\begin{figure*}
  \centering
  \includegraphics[width=0.9\textwidth]{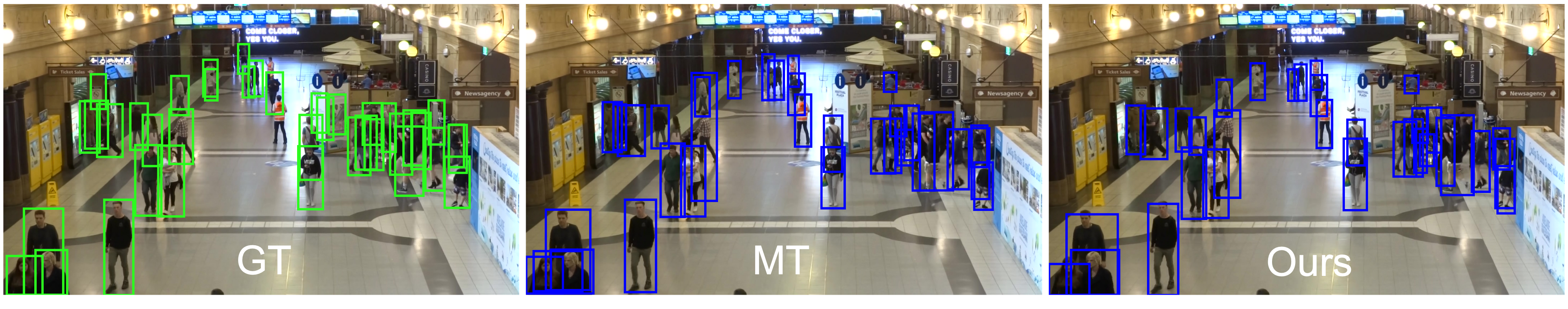}
  \caption{\textbf{Viewpoint Adaptation: Qualitative Results.} We visualize results for viewpoint adaptation between Cityscapes and MOT20-02.
  The left image depicts the ground truth, the middle one the results of Mean Teacher adaptation, and the right one those of our approach. Our approach recovers more detections (e.g., the woman near the stroller in the center-left) while having fewer false positives (overlapping box in bottom-left corner of the MT results).}
  \label{fig:sota_visu_mot}
\end{figure*}

%% file: tables/supp_aug.tex
% !TEX root = ../supp.tex
% !TEX spellcheck = en-US
\begin{table}[h]

  \centering
  \begin{tabular}{lc}
    \toprule

    Kind     & Details  \\
    \midrule
    Random Crop & Relative Range: $[0.3 ,1]$ \\

    Color Jitter & Brightness=$.5$, Hue=$.3$\\
    \bottomrule
  \end{tabular}
  \caption{Augmentations}
  \label{tab:supp_aug}
\end{table}

%% file: figures/lamda_supp.tex
% !TEX root = ../supp.tex
% !TEX spellcheck = en-US
\begin{figure}[h]
\begin{minipage}{0.5\textwidth}
  \centering
  \includegraphics[width=\textwidth]{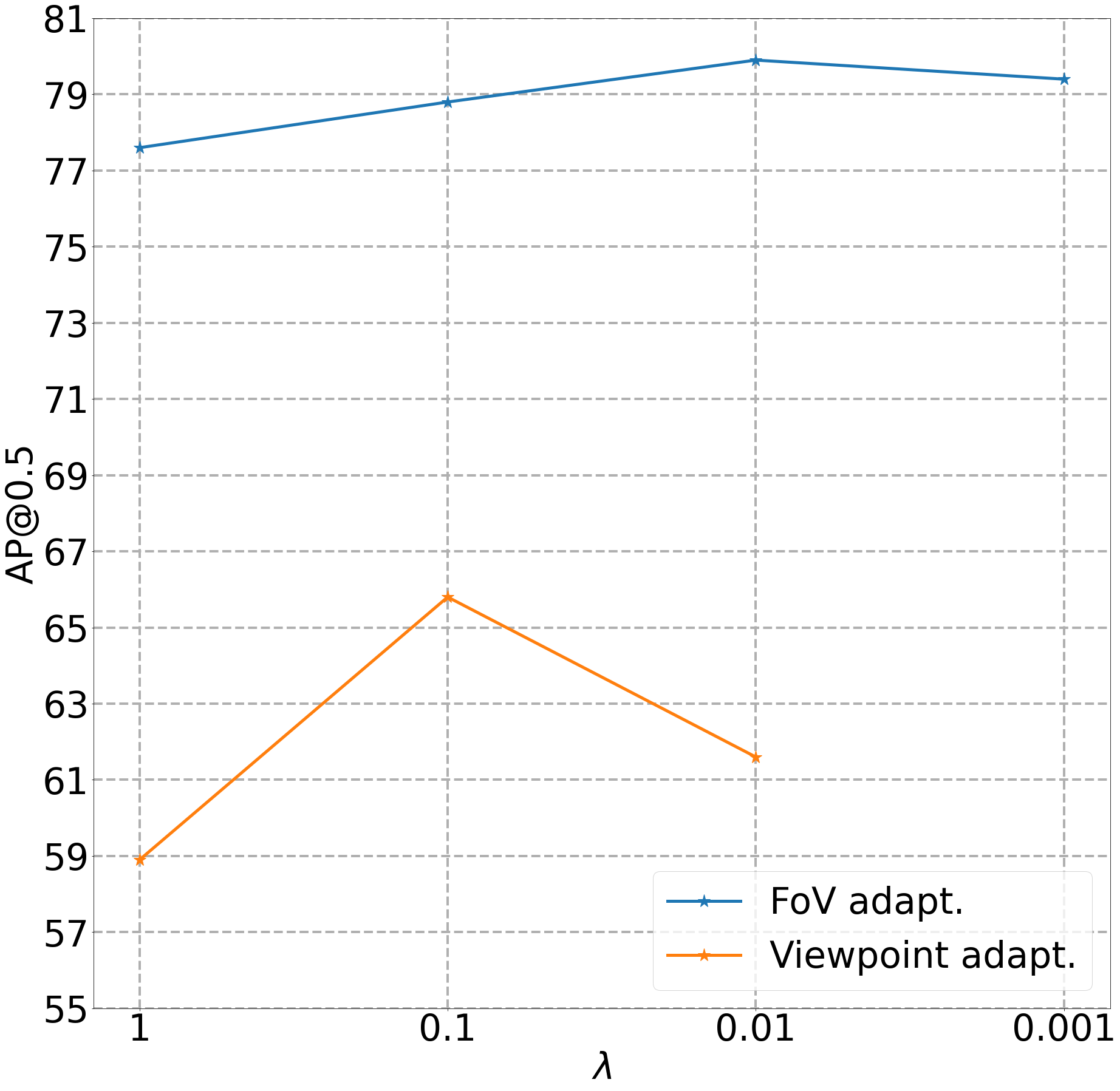}
  \caption{\textbf{Study on $\lambda$} for $\tau=0.6$, $|\mathcal{T}|=5$  }
  \label{fig:supp_lambda}
  \end{minipage}
  \begin{minipage}{0.5\textwidth}
    \centering
  \includegraphics[width=\textwidth]{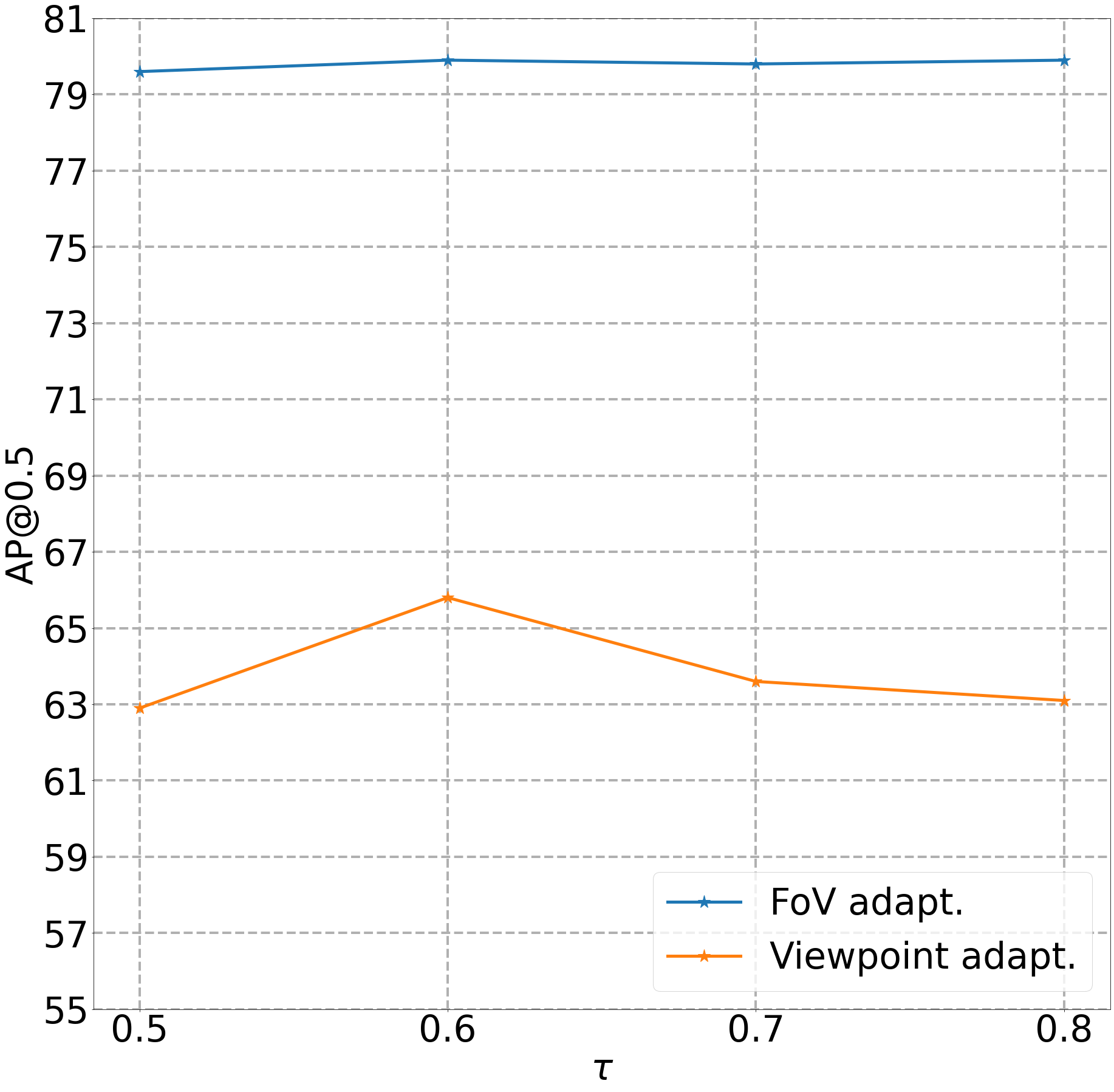}
  \caption{\textbf{Study on $\tau$} for FoV and Viewpoint adaptation using $\lambda=0.01,0.1$, respectively. Here ,$|\mathcal{T}|=5$ is used for the study. }
  \label{fig:supp_tau}
  \end{minipage}
\end{figure}

%% file: figures/tau_supp.tex
% !TEX root = ../supp.tex
% !TEX spellcheck = en-US
% \begin{figure}[t]
%   \centering
%   \includegraphics[width=0.5\textwidth]{figures/tau.png}
%   \caption{\textbf{Study on $\tau$} for FoV and Viewpoint adaptation using $\lambda=0.01,0.1$, respectively. Here ,$|\mathcal{T}|=5$ is used for the study. }
%   \label{fig:supp_tau}
% \end{figure}

%% file: tables/supp_arch_details.tex
% !TEX root = ../supp.tex
% !TEX spellcheck = en-US
\begin{table}[h]
  \caption{Aggregator Architecture for $ | \mathcal{T}|=N$ }
  \label{tab:supp_aggregator}
  \centering
  \begin{tabular}{lcc}
    \toprule
     \multicolumn{2}{r}{ \# Channels}                   \\
    \cmidrule(r){2-3}
    Layer     & Input & Output \\
    \midrule
    Conv2d  $3\times 3 $ &  $ N\times C$ & $ N\times C/2 $ \\
    BatchNorm +  Relu& $ N\times C/2$ & $ N\times C/2$ \\

    Conv2d  $3\times 3 $ &  $ N\times C/2$ & $ C $ \\
    BatchNorm +  Relu & $ C$ & $ C$ \\
    
    Conv2d $1\times 1 $ &  $C$ & $C $ \\
    BatchNorm +  Relu & $ C$ & $ C$ \\
    \bottomrule
  \end{tabular}
\end{table}

%% file: main.bbl
\begin{thebibliography}{10}\itemsep=-1pt

\bibitem{bochkovskiy2020yolov4}
Alexey Bochkovskiy, Chien-Yao Wang, and Hong-Yuan~Mark Liao.
\newblock Yolov4: Optimal speed and accuracy of object detection.
\newblock {\em arXiv preprint arXiv:2004.10934}, 2020.

\bibitem{detr}
Nicolas Carion, Francisco Massa, Gabriel Synnaeve, Nicolas Usunier, Alexander
  Kirillov, and Sergey Zagoruyko.
\newblock Detr, \url{https://github.com/facebookresearch/detr}, 2020.

\bibitem{chen2020harmonizing}
Chaoqi Chen, Zebiao Zheng, Xinghao Ding, Yue Huang, and Qi Dou.
\newblock Harmonizing transferability and discriminability for adapting object
  detectors.
\newblock In {\em Proceedings of the IEEE/CVF Conference on Computer Vision and
  Pattern Recognition}, pages 8869--8878, 2020.

\bibitem{CHEN_2021_I3NET}
Chaoqi Chen, Zebiao Zheng, Yue Huang, Xinghao Ding, and Yizhou Yu.
\newblock I3net: Implicit instance-invariant network for adapting one-stage
  object detectors.
\newblock In {\em IEEE Conference on Computer Vision and Pattern Recognition
  (CVPR)}, 2021.

\bibitem{chen2018domain}
Yuhua Chen, Wen Li, Christos Sakaridis, Dengxin Dai, and Luc Van~Gool.
\newblock Domain adaptive faster r-cnn for object detection in the wild.
\newblock In {\em Proceedings of the IEEE conference on computer vision and
  pattern recognition}, pages 3339--3348, 2018.

\bibitem{cordts2016cityscapes}
Marius Cordts, Mohamed Omran, Sebastian Ramos, Timo Rehfeld, Markus Enzweiler,
  Rodrigo Benenson, Uwe Franke, Stefan Roth, and Bernt Schiele.
\newblock The cityscapes dataset for semantic urban scene understanding.
\newblock In {\em Proceedings of the IEEE conference on computer vision and
  pattern recognition}, pages 3213--3223, 2016.

\bibitem{dai2017deformable}
Jifeng Dai, Haozhi Qi, Yuwen Xiong, Yi Li, Guodong Zhang, Han Hu, and Yichen
  Wei.
\newblock Deformable convolutional networks.
\newblock In {\em Proceedings of the IEEE international conference on computer
  vision}, pages 764--773, 2017.

\bibitem{dendorfer2020mot20}
Patrick Dendorfer, Hamid Rezatofighi, Anton Milan, Javen Shi, Daniel Cremers,
  Ian Reid, Stefan Roth, Konrad Schindler, and Laura Leal-Taix{\'e}.
\newblock Mot20: A benchmark for multi object tracking in crowded scenes.
\newblock {\em arXiv preprint arXiv:2003.09003}, 2020.

\bibitem{deng2021unbiased}
Jinhong Deng, Wen Li, Yuhua Chen, and Lixin Duan.
\newblock Unbiased mean teacher for cross-domain object detection.
\newblock In {\em Proceedings of the IEEE/CVF Conference on Computer Vision and
  Pattern Recognition}, pages 4091--4101, 2021.

\bibitem{everingham2010pascal}
Mark Everingham, Luc Van~Gool, Christopher~KI Williams, John Winn, and Andrew
  Zisserman.
\newblock The pascal visual object classes (voc) challenge.
\newblock {\em International journal of computer vision}, 88(2):303--338, 2010.

\bibitem{ganin2016domain}
Yaroslav Ganin, Evgeniya Ustinova, Hana Ajakan, Pascal Germain, Hugo
  Larochelle, Fran{\c{c}}ois Laviolette, Mario Marchand, and Victor Lempitsky.
\newblock Domain-adversarial training of neural networks.
\newblock {\em The journal of machine learning research}, 17(1):2096--2030,
  2016.

\bibitem{geiger2012we}
Andreas Geiger, Philip Lenz, and Raquel Urtasun.
\newblock Are we ready for autonomous driving? the kitti vision benchmark
  suite.
\newblock In {\em 2012 IEEE Conference on Computer Vision and Pattern
  Recognition}, pages 3354--3361. IEEE, 2012.

\bibitem{gu2021pit}
Qiqi Gu, Qianyu Zhou, Minghao Xu, Zhengyang Feng, Guangliang Cheng, Xuequan Lu,
  Jianping Shi, and Lizhuang Ma.
\newblock Pit: Position-invariant transform for cross-fov domain adaptation.
\newblock In {\em Proceedings of the IEEE/CVF International Conference on
  Computer Vision}, pages 8761--8770, 2021.

\bibitem{he2016deep}
Kaiming He, Xiangyu Zhang, Shaoqing Ren, and Jian Sun.
\newblock Deep residual learning for image recognition.
\newblock In {\em Proceedings of the IEEE conference on computer vision and
  pattern recognition}, pages 770--778, 2016.

\bibitem{inoue2018cross}
Naoto Inoue, Ryosuke Furuta, Toshihiko Yamasaki, and Kiyoharu Aizawa.
\newblock Cross-domain weakly-supervised object detection through progressive
  domain adaptation.
\newblock In {\em Proceedings of the IEEE conference on computer vision and
  pattern recognition}, pages 5001--5009, 2018.

\bibitem{jaderberg2015spatial}
Max Jaderberg, Karen Simonyan, Andrew Zisserman, et~al.
\newblock Spatial transformer networks.
\newblock {\em Advances in neural information processing systems}, 28, 2015.

\bibitem{glenn_jocher_2020_4154370}
Glenn Jocher, Alex Stoken, Jirka Borovec, NanoCode012, ChristopherSTAN, Liu
  Changyu, Laughing, tkianai, Adam Hogan, lorenzomammana, yxNONG, AlexWang1900,
  Laurentiu Diaconu, Marc, wanghaoyang0106, ml5ah, Doug, Francisco Ingham,
  Frederik, Guilhen, Hatovix, Jake Poznanski, Jiacong Fang, Lijun Yu,
  changyu98, Mingyu Wang, Naman Gupta, Osama Akhtar, PetrDvoracek, and Prashant
  Rai.
\newblock {ultralytics/yolov5: v3.1 - Bug Fixes and Performance Improvements},
  Oct. 2020.

\bibitem{kim2019self}
Seunghyeon Kim, Jaehoon Choi, Taekyung Kim, and Changick Kim.
\newblock Self-training and adversarial background regularization for
  unsupervised domain adaptive one-stage object detection.
\newblock In {\em Proceedings of the IEEE/CVF International Conference on
  Computer Vision}, pages 6092--6101, 2019.

\bibitem{kingma2014adam}
Diederik~P Kingma and Jimmy Ba.
\newblock Adam: A method for stochastic optimization.
\newblock {\em arXiv preprint arXiv:1412.6980}, 2014.

\bibitem{li2022cross}
Yu-Jhe Li, Xiaoliang Dai, Chih-Yao Ma, Yen-Cheng Liu, Kan Chen, Bichen Wu,
  Zijian He, Kris Kitani, and Peter Vajda.
\newblock Cross-domain adaptive teacher for object detection.
\newblock In {\em IEEE Conference on Computer Vision and Pattern Recognition
  (CVPR)}, 2022.

\bibitem{long2015learning}
Mingsheng Long, Yue Cao, Jianmin Wang, and Michael Jordan.
\newblock Learning transferable features with deep adaptation networks.
\newblock In {\em International conference on machine learning}, pages 97--105.
  PMLR, 2015.

\bibitem{pei2018multi}
Zhongyi Pei, Zhangjie Cao, Mingsheng Long, and Jianmin Wang.
\newblock Multi-adversarial domain adaptation.
\newblock In {\em Thirty-second AAAI conference on artificial intelligence},
  2018.

\bibitem{redmon2016you}
Joseph Redmon, Santosh Divvala, Ross Girshick, and Ali Farhadi.
\newblock You only look once: Unified, real-time object detection.
\newblock In {\em Proceedings of the IEEE conference on computer vision and
  pattern recognition}, pages 779--788, 2016.

\bibitem{ren2016faster}
Shaoqing Ren, Kaiming He, Ross Girshick, and Jian Sun.
\newblock Faster r-cnn: towards real-time object detection with region proposal
  networks.
\newblock {\em IEEE transactions on pattern analysis and machine intelligence},
  39(6):1137--1149, 2016.

\bibitem{ILSVRC15}
Olga Russakovsky, Jia Deng, Hao Su, Jonathan Krause, Sanjeev Satheesh, Sean Ma,
  Zhiheng Huang, Andrej Karpathy, Aditya Khosla, Michael Bernstein,
  Alexander~C. Berg, and Li Fei-Fei.
\newblock {ImageNet Large Scale Visual Recognition Challenge}.
\newblock {\em International Journal of Computer Vision (IJCV)},
  115(3):211--252, 2015.

\bibitem{saito2019strong}
Kuniaki Saito, Yoshitaka Ushiku, Tatsuya Harada, and Kate Saenko.
\newblock Strong-weak distribution alignment for adaptive object detection.
\newblock In {\em Proceedings of the IEEE/CVF Conference on Computer Vision and
  Pattern Recognition}, pages 6956--6965, 2019.

\bibitem{sakaridis2018semantic}
Christos Sakaridis, Dengxin Dai, and Luc Van~Gool.
\newblock Semantic foggy scene understanding with synthetic data.
\newblock {\em International Journal of Computer Vision}, 126(9):973--992,
  2018.

\bibitem{shen2019scl}
Zhiqiang Shen, Harsh Maheshwari, Weichen Yao, and Marios Savvides.
\newblock Scl: Towards accurate domain adaptive object detection via gradient
  detach based stacked complementary losses.
\newblock {\em arXiv preprint arXiv:1911.02559}, 2019.

\bibitem{sohn2020simple}
Kihyuk Sohn, Zizhao Zhang, Chun-Liang Li, Han Zhang, Chen-Yu Lee, and Tomas
  Pfister.
\newblock A simple semi-supervised learning framework for object detection.
\newblock {\em arXiv preprint arXiv:2005.04757}, 2020.

\bibitem{sun2016deep}
Baochen Sun and Kate Saenko.
\newblock Deep coral: Correlation alignment for deep domain adaptation.
\newblock In {\em European conference on computer vision}, pages 443--450.
  Springer, 2016.

\bibitem{tarvainen2017mean}
Antti Tarvainen and Harri Valpola.
\newblock Mean teachers are better role models: Weight-averaged consistency
  targets improve semi-supervised deep learning results.
\newblock {\em Advances in neural information processing systems}, 30, 2017.

\bibitem{tzeng2017adversarial}
Eric Tzeng, Judy Hoffman, Kate Saenko, and Trevor Darrell.
\newblock Adversarial discriminative domain adaptation.
\newblock In {\em Proceedings of the IEEE conference on computer vision and
  pattern recognition}, pages 7167--7176, 2017.

\bibitem{vs2021mega}
Vibashan VS, Vikram Gupta, Poojan Oza, Vishwanath~A Sindagi, and Vishal~M
  Patel.
\newblock Mega-cda: Memory guided attention for category-aware unsupervised
  domain adaptive object detection.
\newblock In {\em Proceedings of the IEEE/CVF Conference on Computer Vision and
  Pattern Recognition}, pages 4516--4526, 2021.

\bibitem{wu2019detectron2}
Yuxin Wu, Alexander Kirillov, Francisco Massa, Wan-Yen Lo, and Ross Girshick.
\newblock Detectron2.
\newblock \url{https://github.com/facebookresearch/detectron2}, 2019.

\bibitem{xie2018learning}
Shaoan Xie, Zibin Zheng, Liang Chen, and Chuan Chen.
\newblock Learning semantic representations for unsupervised domain adaptation.
\newblock In {\em International conference on machine learning}, pages
  5423--5432. PMLR, 2018.

\bibitem{xu2020adversarial}
Minghao Xu, Jian Zhang, Bingbing Ni, Teng Li, Chengjie Wang, Qi Tian, and
  Wenjun Zhang.
\newblock Adversarial domain adaptation with domain mixup.
\newblock In {\em Proceedings of the AAAI Conference on Artificial
  Intelligence}, volume~34, pages 6502--6509, 2020.

\bibitem{zhu2017unpaired}
Jun-Yan Zhu, Taesung Park, Phillip Isola, and Alexei~A Efros.
\newblock Unpaired image-to-image translation using cycle-consistent
  adversarial networks.
\newblock In {\em Proceedings of the IEEE international conference on computer
  vision}, pages 2223--2232, 2017.

\bibitem{zhu2019adapting}
Xinge Zhu, Jiangmiao Pang, Ceyuan Yang, Jianping Shi, and Dahua Lin.
\newblock Adapting object detectors via selective cross-domain alignment.
\newblock In {\em Proceedings of the IEEE/CVF Conference on Computer Vision and
  Pattern Recognition}, pages 687--696, 2019.

\end{thebibliography}


\begin{thebibliography}{1}\itemsep=-1pt

\bibitem{ren2016faster}
Shaoqing Ren, Kaiming He, Ross Girshick, and Jian Sun.
\newblock Faster r-cnn: towards real-time object detection with region proposal
  networks.
\newblock {\em IEEE transactions on pattern analysis and machine intelligence},
  39(6):1137--1149, 2016.

\bibitem{wu2019detectron2}
Yuxin Wu, Alexander Kirillov, Francisco Massa, Wan-Yen Lo, and Ross Girshick.
\newblock Detectron2.
\newblock \url{https://github.com/facebookresearch/detectron2}, 2019.

\end{thebibliography}
